\documentclass{article}

\usepackage{arxiv}

\usepackage[utf8]{inputenc} 
\usepackage[T1]{fontenc}    
\usepackage{hyperref}       
\usepackage{url}            
\usepackage{booktabs}       
\usepackage{amsfonts}       
\usepackage{nicefrac}       
\usepackage{microtype}      
\usepackage{lipsum}		
\usepackage{graphicx}
\usepackage{natbib}
\usepackage{doi}
\usepackage{multirow} 
\usepackage{amsmath}
\usepackage{graphicx}
\usepackage{array}
\usepackage{times}
\usepackage{amsmath,amssymb,booktabs}
\usepackage{latexsym}
\usepackage{algorithm}
\usepackage{algorithmic}
\usepackage{bm}
\usepackage{makecell}
\usepackage{amsfonts}
\usepackage{multirow}
\usepackage{amsmath} 
\usepackage{graphicx}
\usepackage{subcaption}
\title{Learning to Explain: Prototype-Based Surrogate Models for LLM Classification}

\date{} 					

\author{ Bowen Wei \\
	Department of Computer Science\\
	George Mason University\\
	Fairfax, VA 22030 \\
	\texttt{bwei2@gmu.edu} \\
	\And
        Mehrdad Fazli \\
	Department of Computer Science\\
	George Mason University\\
	Fairfax, VA 22030 \\
	\texttt{mfazli@gmu.edu} \\
	\And
        Ziwei Zhu \\
	Department of Computer Science\\
	George Mason University\\
	Fairfax, VA 22030 \\
	\texttt{zzhu20@gmu.edu} \\
}

\begin{document}
\maketitle

\begin{abstract}

Large language models (LLMs) have demonstrated impressive performance on natural language tasks, but their decision-making processes remain largely opaque. Existing explanation methods either suffer from limited faithfulness to the model's reasoning or produce explanations that humans find difficult to understand. To address these challenges, we propose \textbf{ProtoSurE}, a novel prototype-based surrogate framework that provides faithful and human-understandable explanations for LLMs. ProtoSurE trains an interpretable-by-design surrogate model that aligns with the target LLM while utilizing sentence-level prototypes as human-understandable concepts. Extensive experiments show that ProtoSurE consistently outperforms SOTA explanation methods across diverse LLMs and datasets. Importantly, ProtoSurE demonstrates strong data efficiency, requiring relatively few training examples to achieve good performance, making it practical for real-world applications. 
\end{abstract}

\section{Introduction}
\label{sec:intro}
Large language models (LLMs) have achieved impressive performance across a broad range of natural language tasks. However, their decision-making processes remain largely opaque. This lack of transparency raises serious concerns in high-stakes domains such as healthcare~\cite{yu2018artificial}, law~\cite{zhong2020does}, and finance~\cite{arner2020fintech}, where accurate and understandable reasoning is essential. Attempts to directly analyze the internal computations of LLMs are often computationally intensive, methodologically fragile, and rarely yield explanations that generalize well or are accessible to human users.

Post-hoc explanation methods -- such as SHAP~\cite{lundberg2017unified}, Integrated Gradients~\cite{sundararajan2017axiomatic}, Occlusion~\cite{zeiler2014visualizing}, and DeepLIFT~\cite{shrikumar2017learning} -- explain models by assigning attribution scores to individual tokens. However, this explanation paradigm often fails to capture the actual reasoning of models~\cite{jacovi2020towards} and produces outputs that are difficult for humans to understand~\cite{spectra2021}. An alternative strategy -- prompting LLMs to generate self-explanations~\cite{madsen2024self} or chain-of-thought reasoning~\cite{wei2022cot} -- can yield fluent justifications, but these are often inadequate for revealing models' true inference processes~\cite{lanham2023measuring}. These highlight two key limitations of existing methods: (1) a lack of faithfulness to the model's actual decision-making, and (2) limited human understandability.

\begin{figure*}[t]
   \centering
   \includegraphics[width=\textwidth]{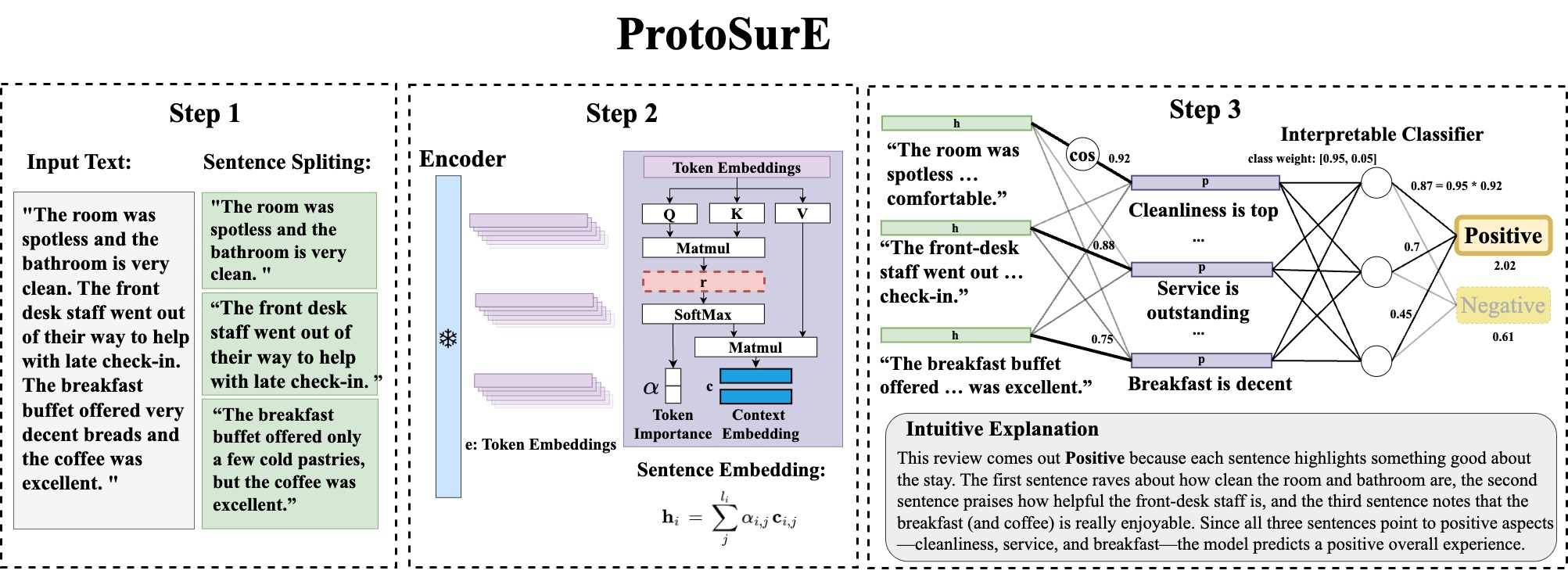}
    \caption{Overview of the ProtoSurE framework. The process consists of three main steps: (1) Sentence splitting that divides input text into semantically meaningful units; (2) An encoder that processes each sentence to generate token embeddings, applying self attention guided by token attribution scores to provide contextualized token embeddings and token attribution weights; and (3) An interpretable classifier that computes cosine similarities between sentence embeddings and learned prototypes, then applies class weights to determine the final prediction. As shown in the explanation box, for a hotel review example, ProtoSurE provides clear rationales by showing which sentences match specific prototypes (Cleanliness, Service, and Breakfast) and how their weighted contributions (0.87, 0.70, and 0.45) lead to the final Positive classification with a score of 2.02 versus 0.61 for Negative.}
    \label{fig:model}
\end{figure*}

To address these limitations, we propose \textbf{ProtoSurE} (\textbf{Proto}type-based \textbf{Sur}rogate \textbf{E}xplanations), a novel framework, as shown in Figure~\ref{fig:model}, that provides faithful and human-understandable explanations for LLM-based text classification. ProtoSurE trains an interpretable-by-design surrogate model to closely approximate the behavior of the target black-box LLM. The surrogate model’s white-box interpretations are then used as faithful explanations of the LLM's predictions. To ensure alignment with the LLM and explanation faithfulness, ProtoSurE employs a knowledge distillation approach~\cite{hinton2015distilling}, training the surrogate to replicate the LLM’s classification behavior.

Moreover, to enhance human comprehension, we design a prototype-based architecture for the surrogate model. Prototype-based methods have proven highly effective and interpretable, making decisions by comparing inputs to learned prototypes that represent meaningful concepts~\cite{chen2019looks}. These approaches have demonstrated strong performance across diverse tasks, including recognition~\cite{chen2019looks}, classification~\cite{li2018deep}, out-of-distribution detection~\cite{ming2019interpretable}, domain adaptation~\cite{tan2018feature}, and segmentation~\cite{donnelly2018deep}. Their intuitive "this looks like that" explanations~\cite{li2018deep} facilitate understanding of complex decisions by linking inputs to human-understandable patterns. In ProtoSurE, we design a sentence-level prototype-based architecture for the surrogate model, producing explanations that align more naturally with how humans understand and reason about language.

ProtoSurE's explanations excel in two critical dimensions: (1) \textbf{faithfulness}, through its rigorous knowledge distillation-based training that ensures the surrogate model accurately reflects the target LLM's behavior; and (2) \textbf{human understandability}, via clear and coherent sentence-level prototypes that align with how humans process language.

In summary, our contributions are: (1) We propose \textbf{ProtoSurE}, a prototype-based surrogate explanation framework for explaining black-box LLM predictions in text classification. (2) We introduce \textbf{sentence-level prototype explanations} that enhance human comprehension significantly beyond existing token-level approaches. (3) We validate ProtoSurE through extensive experiments across diverse text classification benchmarks, demonstrating its effectiveness in delivering faithful and comprehensible explanations.

\section{Related Work}

\paragraph{Post-hoc Explanation Methods.} Post-hoc methods interpret black-box models by revealing input-output relationships. Feature attribution techniques like SHAP~\cite{lundberg2017unified}, Integrated Gradients~\cite{sundararajan2017axiomatic}, and DeepLIFT~\cite{shrikumar2017learning} assign attribution scores to individual tokens. However, these token-level methods struggle with faithfulness~\cite{jacovi2020towards} and human interpretability~\cite{spectra2021}. LLM self-explanations and chain-of-thought reasoning~\cite{wei2022cot}, while promising, produce plausible but unfaithful explanations~\cite{madsen2024self, lanham2023measuring}, limiting their reliability for interpretability.

\paragraph{Prototype-based Neural Networks.} 
Prototype-based methods improve interpretability by comparing inputs with representative examples rather than using abstract feature weights. Originally developed for computer vision~\cite{chen2019looks}, these methods have enabled intuitive "this looks like that" explanations across various applications. Recent adaptations in NLP -- such as PoetryNet~\cite{hong2023protorynet} and ProtoLens~\cite{wei2024advancinginterpretabilitytextclassification} -- have demonstrated effectiveness in delivering white-box text classification. However, these works primarily focus on building interpretable-by-design classifiers, typically based on traditional language models such as BERT~\cite{devlin2019bertpretrainingdeepbidirectional}. In contrast, our work pursues a different goal -- developing a prototype-based model to explain the prediction of a target LLM in a post-hoc way.


\paragraph{Surrogate Models as Explanations.} 
Surrogate models explain complex models by approximating their behavior with simpler, interpretable ones~\cite{ribeiro2016why, molnar2019}. LIME pioneered this approach with local linear approximations, while knowledge distillation techniques~\cite{hinton2015distilling, tan2018distill} transfer complex model knowledge to simpler structures. However, current surrogate methods often face fidelity-interpretability trade-offs~\cite{rudin2019}, especially with LLMs. Our work addresses this challenge by introducing a prototype-based surrogate framework specifically designed to balance faithful approximation of LLM predictions with human-interpretable explanations at the sentence level.

\section{Method}
ProtoSurE learns an interpretable-by-design surrogate model that faithfully explains a target black-box LLM through sentence-level prototype-based explanations. An overview of the model architecture is shown in Figure~\ref{fig:model}.

\subsection{Overall Structure}
\noindent\textbf{Problem Formulation.} Given a target black-box LLM $\mathcal{M}_{\mathrm{target}}$ and a set of text samples with corresponding predictions from $\mathcal{M}_{\mathrm{target}}$, our goal is to construct an interpretable surrogate model that faithfully explains these predictions. Specifically, the set of text samples is denoted as $\mathcal{X} = \{X_1, X_2, \dots, X_N\}$. The corresponding predictions by $\mathcal{M}_{\mathrm{target}}$ are denoted as $\hat{\mathcal{Y}} = \{\hat{y}_1, \hat{y}_2, \dots, \hat{y}_N\}$. 

To increase alignment between our surrogate model and $\mathcal{M}_{\mathrm{target}}$, we leverage token-level attribution scores $\{\mathbf{r}_1, \mathbf{r}_2, \dots, \mathbf{r}_N\}$ obtained from any existing post-hoc explanation methods~\cite{chefer2021transformerinterpretabilityattentionvisualization} applied to $\mathcal{M}_{\mathrm{target}}$. These scores provide guidance about which tokens are most influential in the $\mathcal{M}_{\mathrm{target}}$'s decision process, helping the surrogate model focus on the same textual elements that drive the LLM's predictions. Experiments in Section~\ref{sec:token_attr} confirm the effectiveness of incorporating these attribution scores.


\noindent \textbf{Model Overview.} As illustrated in Figure~\ref{fig:model}, ProtoSurE processes input text through three main steps to generate explanations for LLM classifications. 

In Step 1, ProtoSurE splits the input text into sentences using standard punctuation delimiters (., !, ?). Figure~\ref{fig:model} shows a movie review segmented into three distinct sentences.


In Step 2, each sentence is first tokenized and passed through a text encoder to generate token embeddings. These token embeddings are then processed through a self-attention module to determine token importance $\alpha$ and create contextual embeddings $\mathbf{c}$ for tokens. The sentence embedding is computed as a weighted average of the contextual token embeddings, with weights determined by token attribution scores ($h_i = \sum_j \alpha_{i,j} \mathbf{c}_{i,j}$). 

In Step 3, the sentence embeddings are compared against a set of trainable prototypes $\mathcal{P} = \{\mathbf{p}_k \in \mathbb{R}^d : k = 1, \dots, K\}$, where each prototype is represented by an embedding vector, and the hyperparameter $K$ is the number of prototypes specified. Each sentence activates different prototypes based on semantic similarity, with an interpretable classification head (such as a logistic regression model or decision tree) taking these activations as inputs to determine the final prediction.

\noindent\textbf{Explanation Generation.} Figure~\ref{fig:model} illustrates this process on a hotel review classified as Positive. To simplify the example, only the most important prototype for each sentence is shown (each sentence still has nonzero similarities to other prototypes, but we show the largest contributor in this example). The first sentence (``The room was spotless and the bed was extremely comfortable.'') activates the \textit{Cleanliness} prototype (cosine similarity = 0.92) with a learned positive-class weight of 0.95, contributing 0.92 × 0.95 = 0.87 to the positive logit; the second sentence (``The front-desk staff went out of their way to help with late check-in.'') activates the \textit{Service} prototype (similarity = 0.88) with weight 0.80, contributing 0.70; and the third sentence (``The breakfast buffet offered only a few cold pastries, but the coffee was excellent.'') activates the \textit{Breakfast} prototype (similarity = 0.75) with weight 0.60, contributing 0.45. These three main contributions sum to a positive-class score of 2.02 versus a negative-class score of 0.45, yielding the \textbf{Positive} prediction. By grounding each sentence in the most influential prototypes, this framework delivers explanations that are both faithful to the model's internal reasoning and easy to understand.

\subsection{Attribution-aware Sentence Embedding}
A key innovation in ProtoSurE is its ability to create sentence embeddings that capture both semantic meaning and relevance to the classification task. 

Given an input text $X$, we first segment it into sentences $\mathbf{s} = [s_1, s_2, \dots, s_M]$. Each sentence $s_i$ is tokenized into $\mathbf{t}^{(i)} = [t_{i,1}, \dots, t_{i,\ell_i}]$ and encoded using a pre-trained text encoder $\mathcal{E}$ (e.g., MPNet~\cite{song2020mpnet}, BGE~\cite{chen2024bge}), yielding token embeddings:
\begin{equation}
\mathbf{e}_{i,j} = \mathcal{E}(t_{i,j}) \in \mathbb{R}^d.
\end{equation}

To ensure the learned surrogate model behaves similarly to the target LLM, we aim to incorporate the information about which tokens the LLM relies on for prediction. We can use established post-hoc explanation methods (e.g., attention relevancy maps~\cite{chefer2021transformerinterpretabilityattentionvisualization}, integrated gradients~\cite{sundararajan2017axiomatic}, or SHAP~\cite{lundberg2017unified}) to obtain attribution scores $r_{i,j}$ for each token. These methods quantify the contribution of individual tokens to the model's predictions. We normalize the token attribution scores $r_{i,j}$ obtained from the target LLM:
\begin{equation}
\hat{r}_{i,j} = \frac{r_{i,j}}{\sum_{j'=1}^{\ell_i} r_{i,j'} + \epsilon}, \quad \epsilon = 10^{-9}.
\end{equation}

We then use these normalized attribution scores to guide a self-attention mechanism. With query, key, and value matrices $\mathbf{Q}_i, \mathbf{K}_i, \mathbf{V}_i \in \mathbb{R}^{\ell_i \times d}$ derived from token embeddings, we compute attention as:
\begin{equation}
A_i = \mathrm{softmax}\left(\frac{\mathbf{Q}_i \mathbf{K}_i^\top}{\sqrt{d}} + \mathbf{\hat{r}}_i\right),
\end{equation}
\begin{equation}
\mathbf{c}_i = A_i \mathbf{V}_i.
\end{equation}

To quantify the importance of each token within a sentence, we compute token-level attribution scores $\alpha_{i,j}$. In self-attention, each token position attends to all token positions, creating an attention matrix $A$ where $A_{i,k,j}$ represents the attention weight from token $k$ to token $j$ in sentence $i$. We average these attention weights across all source positions to obtain a measure of how much attention each token receives:
\begin{equation}
\alpha_{i,j} = \frac{\exp\left(\frac{1}{\ell_i} \sum_{k=1}^{\ell_i} A_{i,k,j}\right)}{\sum_{j'=1}^{\ell_i} \exp\left(\frac{1}{\ell_i} \sum_{k=1}^{\ell_i} A_{i,k,j'}\right)}.
\end{equation}
These attribution scores are then used to create attribution-aware sentence embeddings through weighted pooling:
\begin{equation}
\mathbf{h}_i = \sum_{j=1}^{\ell_i} \alpha_{i,j} \cdot \mathbf{c}_{i,j} \in \mathbb{R}^d,
\end{equation}

This approach ensures that our sentence embeddings not only represent semantic content but also reflect the relative importance of different parts of the text to the target LLM's classification decision.

\subsection{Prototype Learning and Classification}
To create meaningful and diverse prototypes that represent patterns in the data, we initialize prototype embeddings using a clustering-based approach. We first encode all sentences from the training data into embeddings. We then apply K-means clustering to these embeddings and use the resulting cluster centers as initial prototype embeddings. To provide interpretable explanations, we associate each prototype with its nearest sentence from the training data based on cosine similarity. The prototype embeddings can either be fixed after initialization or further refined through training. We conducted experiments to compare the performance of these two settings in Section~\ref{sec:p_update}.

For classification, each sentence embedding $\mathbf{h}_i$ is compared to all prototype embeddings using cosine similarity: $a_{i,k} = \cos(\mathbf{h}_i, \mathbf{p}_k)$. This produces a similarity vector $\mathbf{a}_i = [a_{i,1}, a_{i,2}, \dots, a_{i,P}]$ that represents how strongly each sentence activates each prototype. We then apply a linear classifier to this similarity vector to generate prediction logits for each sentence: $\tilde{y}_i = f(\mathbf{a}_i)$. The final prediction for the entire text sample is computed by summing up these sentence-level predictions.

This approach ensures transparency in the classification process by enabling us to trace how each input sentence contributes to the final prediction based on its similarity to specific prototypes. The predictions can then be explained in human-understandable terms by showing how each sentence aligns with meaningful prototypical patterns.

\subsection{Training Objective}

ProtoSurE is trained with a multi-objective loss balancing fidelity, prototype coverage, and diversity:
\begin{equation}
\mathcal{L} = \text{CrossEntropy}(\hat{y}, \tilde{y}) + \lambda_1 \mathcal{L}_{\text{proto}} + \lambda_2 \mathcal{L}_{\text{diversity}}.
\end{equation}

Hyperparameters $\lambda_1$ and $\lambda_2$ are set to 0.1 in our experiments. The prototype utilization loss encourages each prototype to match at least one training sentence:
\begin{equation}
\mathcal{L}_{\text{proto}} = -\frac{1}{P} \sum_{k=1}^P \max_{i} \text{sim}(\mathbf{h}_i, \mathbf{p}_k).
\end{equation}

The diversity loss penalizes overlap between prototypes:
\begin{equation}
\mathcal{L}_{\text{diversity}} = \frac{1}{P(P-1)} \sum_{i=1}^P \sum_{\substack{j=1 \\ j \neq i}}^P |\mathbf{p}_i^\top \mathbf{p}_j|.
\end{equation}

\section{Experiments}

In this section, we evaluate ProtoSurE across multiple dimensions to answer the following research questions: \textbf{RQ1:} How faithfully does ProtoSurE explain LLM predictions compared to existing explanation methods? \textbf{RQ2: } How does the training data size affect ProtoSurE's performance? \textbf{RQ3:} How do core components (encoder, token attribution score, prototype updating) contribute to model effectiveness? \textbf{RQ4:} How do key hyperparameters influence performance? \textbf{RQ5:} How do ProtoSurE explanations look like in a case study?

\subsection{Experimental Setup}

\noindent \textbf{Datasets.} We evaluate ProtoSurE on four diverse text classification datasets spanning single-label, multi-label, and domain-specific classification tasks: IMDB, Hotel, DBPedia, and Consumer Complaint. Details are provided in Appendix~\ref{appendix:datasets}.

\noindent\textbf{Reproducibility.} ProtoSurE was implemented using PyTorch. We train our model with the following hyperparameters: learning rate selected from \{1e-2, 2e-2, 2e-3\} with AdamW optimizer \cite{loshchilov2017decoupled}, batch size of 16, and training for 10 epochs. The prototype number ($P$) is selected from \{10, 20, 40, 100\}, and we set $\lambda_1 = 0.1$, and $\lambda_2 = 0.1$ for the loss components. We employ the relevancy map approach proposed by \citet{chefer2021transformerinterpretabilityattentionvisualization} as the token attribution scores, which propagates classification-relevant gradients through the attention layers to identify important tokens in the LLM's decision process. The experiments were conducted on NVIDIA A100 80GB GPUs. 

\noindent\textbf{Baselines.} We compare ProtoSurE against four widely-used post-hoc explanation methods: SHAP \cite{lundberg2017unified}, Integrated Gradients (IG) \cite{sundararajan2017axiomatic}, Occlusion \cite{zeiler2014visualizing}, and DeepLIFT \cite{shrikumar2017learning}. Each method is applied to explain predictions from four target LLMs: Llama-3.1-8B-Instruct, Llama-3.2-3B, Qwen2.5-7B-Instruct-1M, and Mistral-7B-Instruct-v0.2. For fair comparison, we adapt all baseline methods to provide sentence-level attributions by aggregating token-level scores. Detailed descriptions of the baseline methods are provided in Appendix \ref{app:baselines}.

\noindent\textbf{Evaluation Metrics.} We assess faithfulness using seven metrics: Accuracy (Acc), Comprehensiveness (Comp) \cite{DeYoung2020}, Sufficiency (Suff) \cite{DeYoung2020}, Decision Flip Fraction (DFF) \cite{serrano2019attention}, Decision Flip with Most Important Sentence (DFS) \cite{chrysostomou2021improving}, Deletion Rank Correlation (Del) \cite{alvarez2018towards}, and Insertion Rank Correlation (Ins) \cite{luss2021leveraging}. Details are provided in Appendix \ref{app:metrics}.

\subsection{Faithfulness Evaluation (RQ1)}

\begin{table}[t]
\centering
\caption{Average ranking of explanation methods across six faithfulness metrics evaluated on all LLMs (Llama-3.1-8B, Llama-3.2-3B, Qwen2.5-7B, Mistral-7B) and all datasets. Lower rank indicates better performance. }
\label{tab:overall_rankings}
\small
\setlength{\tabcolsep}{3pt}
\renewcommand{\arraystretch}{1.1}
\begin{tabular}{lccccccc}
  \toprule
  \textbf{Method} & \textbf{Comp} & \textbf{Suff} & \textbf{DFF} & \textbf{DFS} & \textbf{Del} & \textbf{Ins} & \textbf{Overall} \\
  \midrule
  SHAP      & 3.50 & 4.03 & 3.62 & 3.56 & 4.00 & 2.88 & 3.60 \\
  IG        & 2.19 & 2.09 & 2.28 & 2.81 & 3.09 & 2.28 & 2.46 \\
  Occl      & 3.44 & 2.78 & 3.69 & 2.88 & 2.03 & 3.22 & 3.01 \\
  DeepLIFT  & 4.88 & 4.41 & 4.34 & 4.06 & 4.12 & 4.50 & 4.39 \\
  \textbf{ProtoSurE} & \textbf{1.00} & \textbf{1.69} & \textbf{1.06} & \textbf{1.69} & \textbf{1.75} & \textbf{2.12} & \textbf{1.55} \\
  \bottomrule
\end{tabular}
\end{table}

In this section, we evaluate how faithfully ProtoSurE explains the predictions of target LLMs compared to SOTA post-hoc explanation methods. We assess faithfulness from two perspectives: (1) alignment with LLM behavior through accuracy on predicting classification results of the LLM, and (2) fidelity to the LLM's underlying reasoning process through established faithfulness metrics.

Our comprehensive faithfulness evaluation in Table~\ref{tab:faithfulness_part2} (Appendix~\ref{appendix:faith_eval}) examines performance across four target LLMs and four datasets. First, ProtoSurE demonstrates strong behavioral alignment with target LLMs, achieving an average classification accuracy of 89.58\% across all experiments. This high accuracy is essential for ensuring that our explanations reflect the actual decisions made by the target LLM. Table~\ref{tab:overall_rankings} summarizes the average rankings across all faithfulness metrics. ProtoSurE consistently outperforms baseline methods, achieving the best overall average ranking (1.55) compared to IG (2.46), Occlusion (3.01), SHAP (3.60), and DeepLIFT (4.39). This comprehensive advantage demonstrates our approach's superior ability to faithfully capture LLM reasoning processes.

ProtoSurE consistently outperforms baseline methods across all faithfulness metrics. It achieves the highest Comprehensiveness (Comp) score (0.235 vs. IG's 0.164), with a 55\% gain on Mistral-7B for DBPedia (0.389 vs. 0.250); shows better Sufficiency (Suff) with a lower average (0.131 vs. IG's 0.141), notably on Hotel with Llama-3.1-8B (0.060 vs. 0.114); improves stability in Decision Flip Fraction (DFF) (0.685 vs. 0.706), especially on Mistral-7B for IMDB (0.512 vs. 0.541); outperforms in Decision Flip with Most Important Sentence (DFS) (0.187 vs. 0.158), with strong results on Mistral-7B for IMDB (0.325 vs. 0.285); and leads in both Deletion (Del: 0.105 vs. 0.095) and Insertion Rank Correlation (Ins: 0.336 vs. 0.334).

These gains support our central claim: sentence-level prototypes offer a more natural granularity for explaining LLM behavior, capturing semantic reasoning aligned with human understanding. The benefit is especially evident for larger instruction-tuned models like Llama-3.1-8B, where sentence-level explanations more faithfully reflect the model’s decision process.

\subsection{Impact of Training Data Size (RQ2)}

\begin{figure}[t]
  \centering
  \begin{subfigure}[b]{0.5\textwidth}
    \centering
    \includegraphics[width=\textwidth]{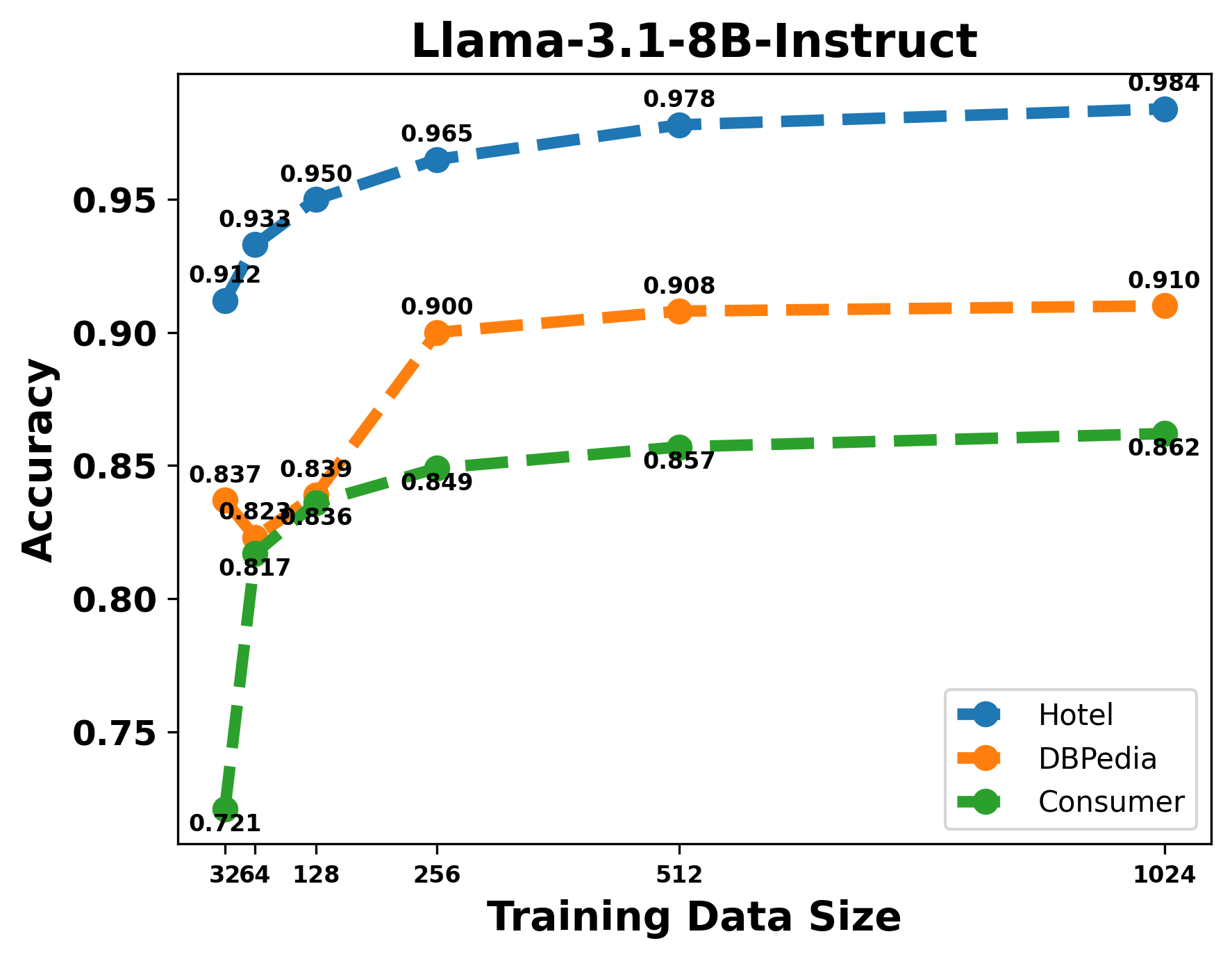}
    \caption{Llama-3.1-8B-Instruct}
  \end{subfigure}\hfill
  \begin{subfigure}[b]{0.5\textwidth}
    \centering
    \includegraphics[width=\textwidth]{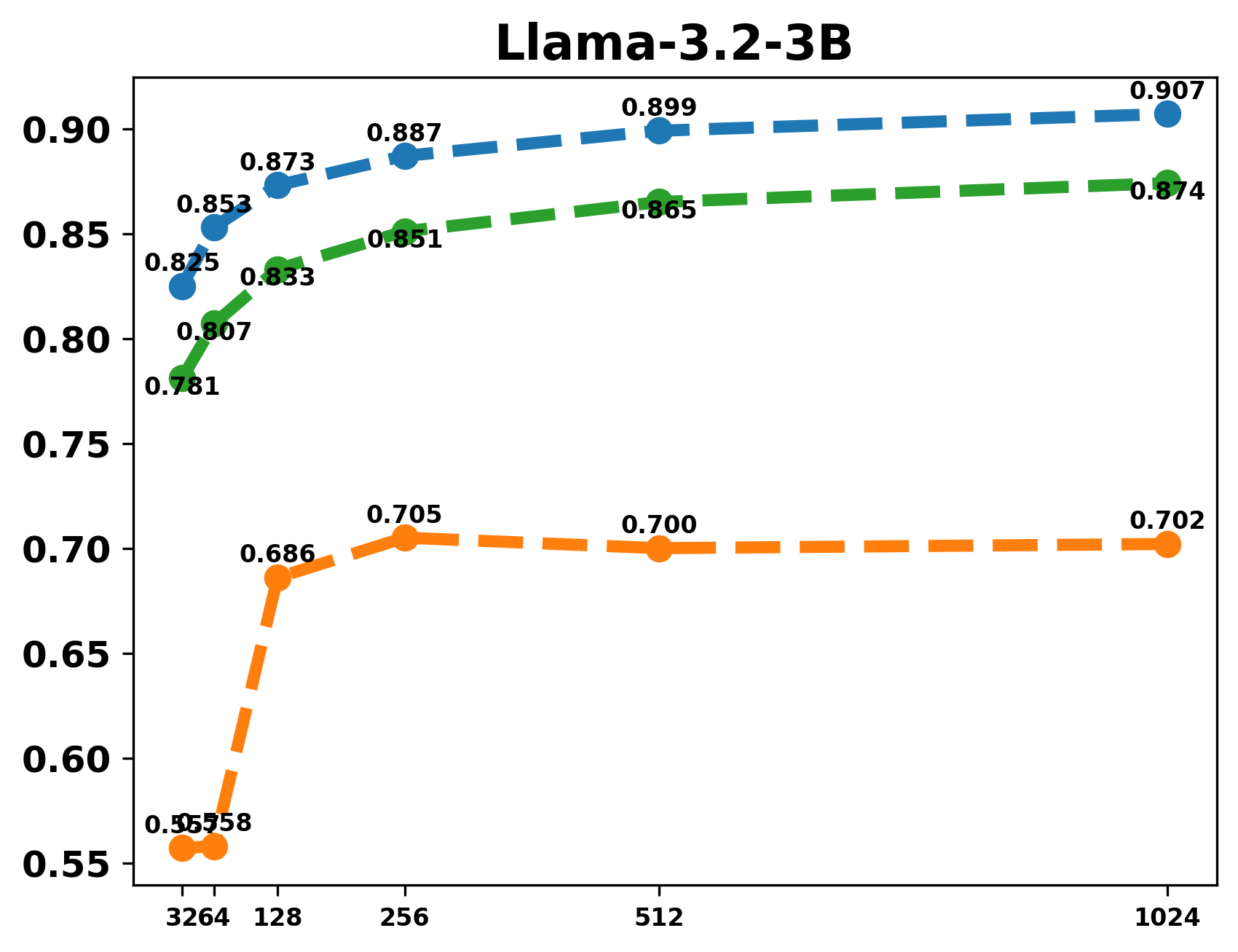}
    \caption{Llama-3.2-3B}
  \end{subfigure}

  \begin{subfigure}[b]{0.5\textwidth}
    \centering
    \includegraphics[width=\textwidth]{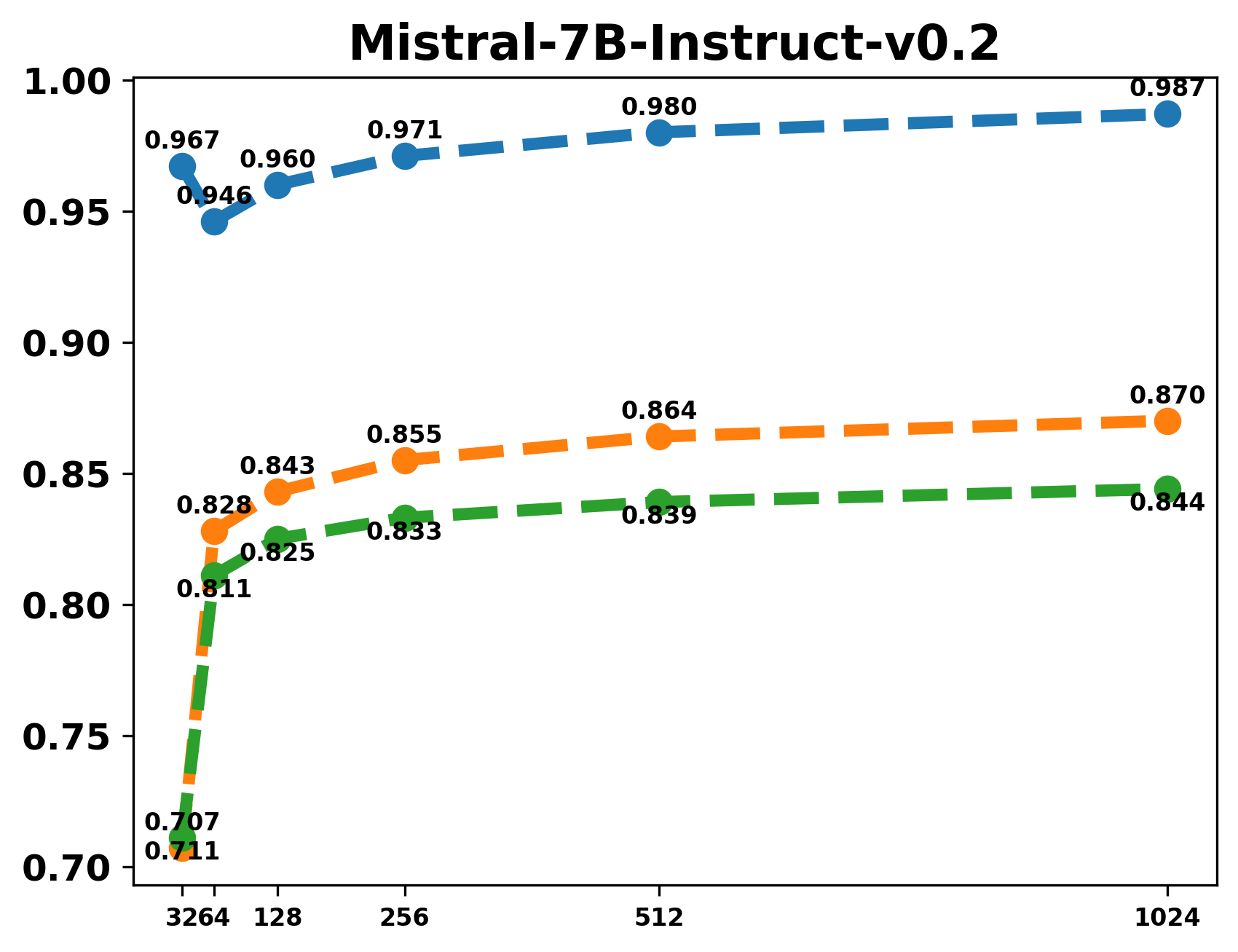}
    \caption{Mistral‑7B‑Instruct‑v0.2}
  \end{subfigure}\hfill
  \begin{subfigure}[b]{0.5\textwidth}
    \centering
    \includegraphics[width=\textwidth]{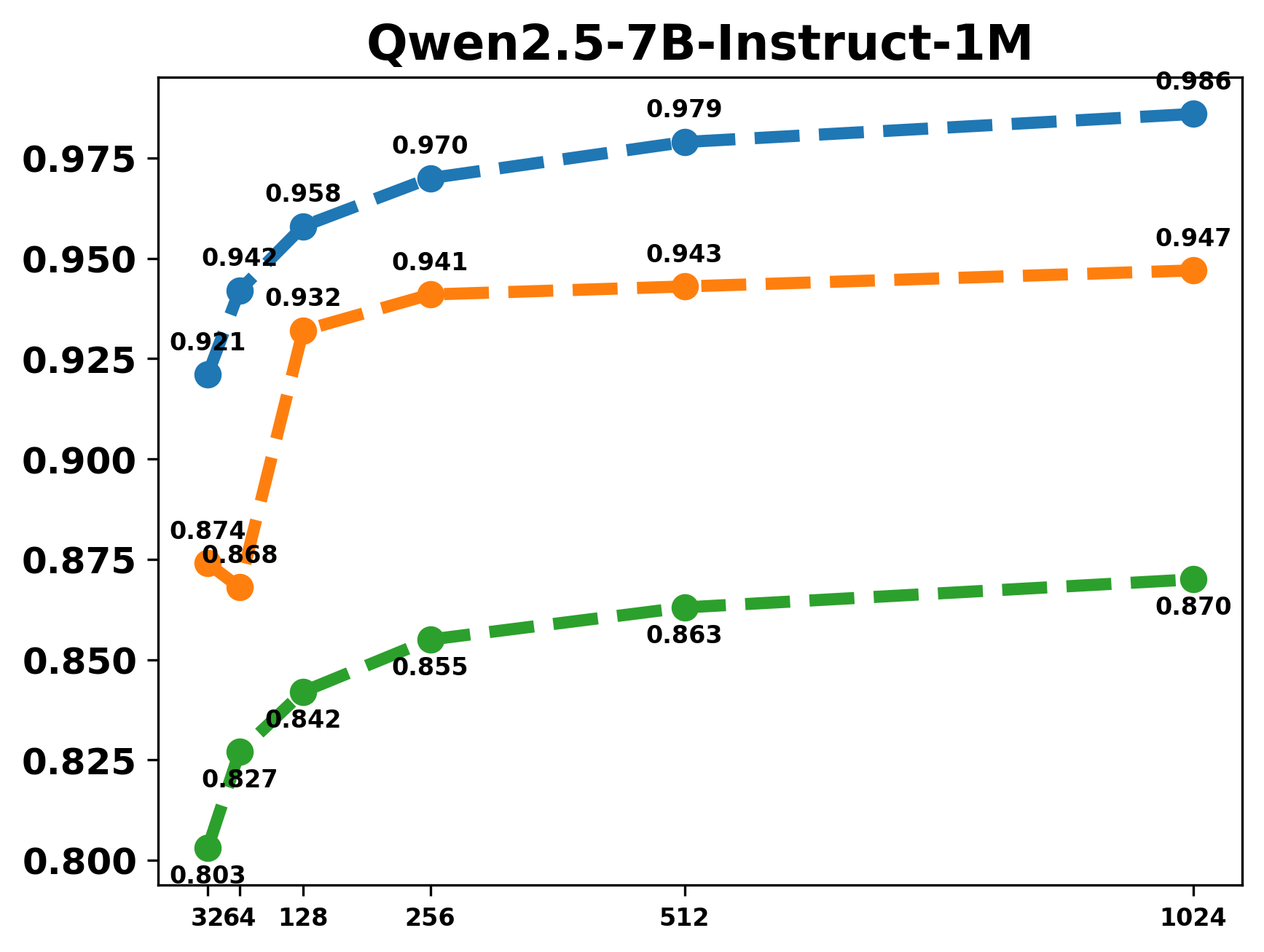}
    \caption{Qwen2.5‑7B‑Instruct‑1M}
  \end{subfigure}

\caption{Impact of training data size on ProtoSurE's accuracy across different datasets and target LLMs. Results show consistent performance improvements as training data increases from 32 to 1024 examples. Notably, ProtoSurE achieves reasonable performance (80-85\% accuracy) with just 128 training examples across most LLMs and datasets.}
\label{fig:data_dependency}
\end{figure}
In this section, we investigate how training data size affects ProtoSurE's ability to faithfully reproduce target LLM predictions. Figure~\ref{fig:data_dependency} presents performance across varying training data sizes for all four target LLMs and three datasets. The key finding is that \textbf{ProtoSurE requires relatively little training data to effectively align with target LLMs}. With just 128 training examples, ProtoSurE achieves strong performance (80-85\% accuracy) across most LLMs and datasets, demonstrating its data efficiency in capturing LLM decision patterns. For Llama-3.1-8B-Instruct on the Hotel dataset, accuracy reaches 95.0\% with just 128 examples, approaching the 98.4\% achieved with 1024 examples. Similarly, for DBPedia classification using Mistral-7B, performance with 128 examples (85.5\%) closely approximates that with 1024 examples (87.0\%). We observe that performance improvements generally plateau after 512 examples, with the Consumer dataset showing more pronounced benefits from additional data across all models. This data efficiency makes ProtoSurE particularly suitable for real-world applications where collecting large labeled datasets maybe impractical.

\subsection{Ablation Study (RQ3)}

We conduct a comprehensive ablation study to understand how different components contribute to ProtoSurE's overall performance, focusing on encoder selection, token importance design, and prototype updating strategies.

\begin{table}[t]
\centering
\caption{Impact of different encoders on ProtoSurE's Accuracy (\%) for Llama-3.1-8B-Instruct across datasets. Full results across all target LLMs are in the Appendix~\ref{appendix: encoder}. Best results are in \textbf{bold}.}
\label{tab:embedding_impact1}
\small
\setlength{\tabcolsep}{2pt}
\renewcommand{\arraystretch}{1.1}
\begin{tabular}{lcccc}
\toprule
\textbf{Encoder} & \textbf{Hotel} & \textbf{DBPedia} & \textbf{Consumer} & \textbf{Avg Rank}  \\
\midrule
\multicolumn{5}{c}{\textit{Llama-3.1-8B-Instruct}} \\
\midrule
SBERT       & 0.970  & 0.907 & 0.859  & 3.33 \\
BGE       & \textbf{0.991}  & 0.906 & \textbf{0.863}  & 2.00 \\
GTE                & 0.984  & \textbf{0.910} & 0.862 & \textbf{1.67} \\
E5   & 0.989  & 0.908 & 0.860 & 2.67 \\
T5        & 0.978  & 0.891 & 0.844 & 5.00 \\

\bottomrule
\end{tabular}
\end{table}

\begin{table}[t]
\centering
\small
\caption{Overall average accuracy and rank of encoders across all target LLMs and three datasets.}
\label{tab:avg_rank_overall_inline}
\renewcommand{\arraystretch}{1.1}
\begin{tabular}{lcc}
\toprule
\textbf{Encoder} & \textbf{Overall Avg} & \textbf{Avg Rank} \\
\midrule
GTE & \textbf{0.8952} & \textbf{2.00} \\
BGE & 0.8939 & 2.50 \\
E5 & 0.8933 & 3.00 \\
SBERT & 0.8881 & 3.75 \\
T5 & 0.8891 & 3.75 \\
\bottomrule
\end{tabular}
\end{table}

\subsubsection{Encoder Impact}
The choice of encoder is an important component of ProtoSurE, as it determines the quality of sentence embeddings. We evaluate five state-of-the-art encoders: SBERT~\cite{reimers-2019-sentence-bert}, BGE~\cite{chen2024bge}, GTE~\cite{li2023towards}, E5~\cite{wang2022text}, and T5~\cite{ni2021sentence}.

Table~\ref{tab:embedding_impact1} presents accuracy results for Llama-3.1-8B-Instruct across three datasets, with full results across all target LLMs available in Appendix~\ref{appendix: encoder}. Notably, all encoders achieve strong performance, with accuracies generally above 0.84 across all datasets and target LLMs. As shown in Table~\ref{tab:avg_rank_overall_inline}, when averaging across all target LLMs and datasets, GTE achieved the highest overall accuracy (0.8952) and average rank (2.00), followed closely by BGE (0.8939, 2.50) and E5 (0.8933, 3.00).

The relatively small performance differences between encoders (within ~0.007 accuracy points) demonstrate that ProtoSurE is robust and not limited to any single embedding model. This flexibility is particularly valuable, as it allows practitioners to select encoders based on specific requirements such as efficiency, domain alignment, or resource constraints without performance degradation.

\subsubsection{Token Attribution Score Impact}
\label{sec:token_attr}
\begin{table}[t]
 \centering
 \caption{Effect of token relevancy maps on accuracy (\%) across all target LLMs and datasets. The relevancy map delivers consistent performance improvements across all models and datasets. Best results are in \textbf{bold}.}
 \label{tab:relevancy_map}
 \small
 \setlength{\tabcolsep}{3.5pt}
 \renewcommand{\arraystretch}{1.1}
 \begin{tabular}{lcccc}
   \toprule
   \textbf{Model Variant} & \textbf{Hotel} & \textbf{DBPedia} & \textbf{Consumer} & \textbf{Avg}  \\
   \midrule
   \multicolumn{5}{c}{\textit{Llama-3.1-8B-Instruct}} \\
   \midrule
   w/o token attribution & 0.975 & 0.901 & 0.856 & 0.911 \\
   \textbf{w/ token attribution} & \textbf{0.984} & \textbf{0.910} & \textbf{0.862} & \textbf{0.919} \\
   \midrule
   \multicolumn{5}{c}{\textit{Llama-3.2-3B}} \\
   \midrule
   w/o token attribution & 0.898 & 0.695 & 0.866 & 0.820 \\
   \textbf{w/ token attribution} & \textbf{0.907} & \textbf{0.702} & \textbf{0.874} & \textbf{0.828} \\
   \midrule
   \multicolumn{5}{c}{\textit{Qwen2.5-7B-Instruct-1M}} \\
   \midrule
   w/o token attribution & 0.979 & 0.939 & 0.863 & 0.927 \\
   \textbf{w/ token attribution} & \textbf{0.986} & \textbf{0.947} & \textbf{0.870} & \textbf{0.934} \\
   \midrule
   \multicolumn{5}{c}{\textit{Mistral-7B-Instruct-v0.2}} \\
   \midrule
   w/o token attribution & 0.980 & 0.863 & 0.838 & 0.894 \\
   \textbf{w/ token attribution} & \textbf{0.987} & \textbf{0.870} & \textbf{0.844} & \textbf{0.900} \\
   \bottomrule
 \end{tabular}
\end{table}

In this section, we examine whether incorporating token-level attributions from the target LLM enhances ProtoSurE's ability to mimic LLM behavior. Table~\ref{tab:relevancy_map} compares ProtoSurE variants with and without token attribution integration across all target LLMs and datasets. The results demonstrate consistent performance improvements when leveraging token attributions, with average accuracy gains ranging from 0.6 to 0.8 percentage points across different LLMs. For instance, incorporating token attribution scores with Llama-3.1-8B improves average accuracy from 91.1\% to 91.9\%, while similar gains are observed with Qwen2.5-7B (92.7\% to 93.4\%) and Llama-3.2-3B (82.0\% to 82.8\%).

These performance gains validate the effectiveness of token-level attributions. By prioritizing tokens deemed significant by the target LLM, ProtoSurE creates more faithful sentence representations that align with the original model's reasoning.

\subsubsection{Prototype Update}
\label{sec:p_update}
\begin{table}[t]
\centering
\caption{Impact of prototype updating strategies on ProtoSurE's Accuracy (\%) across all target LLMs and datasets. Best results for each target LLM-dataset combination are in \textbf{bold}.}
\label{tab:prototype_updating}
\small
\setlength{\tabcolsep}{3.5pt}
\renewcommand{\arraystretch}{1.1}
\begin{tabular}{lcccc}
  \toprule
  \textbf{Update Strategy} & \textbf{Hotel} & \textbf{DBPedia} & \textbf{Consumer} & \textbf{Avg}  \\
  \midrule
  \multicolumn{5}{c}{\textit{Llama-3.1-8B-Instruct}} \\
  \midrule
  w/o update & 97.9 & 89.5 & 81.8 & 89.7 \\
  \textbf{w/ update (Ours)} & \textbf{98.4} & \textbf{91.0} & \textbf{83.2} & \textbf{90.9} \\
  \midrule
  \multicolumn{5}{c}{\textit{Llama-3.2-3B}} \\
  \midrule
  w/o update & 89.3 & 69.8 & 78.6 & 79.2 \\
  \textbf{w/ update (Ours)} & \textbf{90.7} & \textbf{71.7} & \textbf{80.1} & \textbf{80.8} \\
  \midrule
  \multicolumn{5}{c}{\textit{Qwen2.5-7B-Instruct-1M}} \\
  \midrule
  w/o update & 98.5 & 93.8 & 85.6 & 92.6 \\
  \textbf{w/ update (Ours)} & \textbf{99.0} & \textbf{94.7} & \textbf{87.0} & \textbf{93.6} \\
  \midrule
  \multicolumn{5}{c}{\textit{Mistral-7B-Instruct-v0.2}} \\
  \midrule
  w/o update & 98.1 & 86.2 & 83.5 & 89.3 \\
  \textbf{w/ update (Ours)} & \textbf{98.7} & \textbf{87.5} & \textbf{84.6} & \textbf{90.3} \\
  \bottomrule
\end{tabular}
\end{table}

We investigate whether prototype vectors should be updated during training or kept fixed after initialization. Table~\ref{tab:prototype_updating} demonstrates consistent improvements when prototypes are trainable across all target LLMs and datasets. While fixed prototypes captured by clustering provide a reasonable starting point, allowing them to adapt during training enables more refined decision boundaries that better approximate the target LLM's behavior. Importantly, the updated prototypes maintain interpretability while achieving better alignment with the target LLM's classification patterns, offering an optimal balance between accuracy and human-understandable explanations.

\subsection{Hyperparameter Study (RQ4)}
\begin{figure}[t]
  \centering
  \begin{subfigure}[b]{0.5\textwidth}
    \centering
    \includegraphics[width=\textwidth]{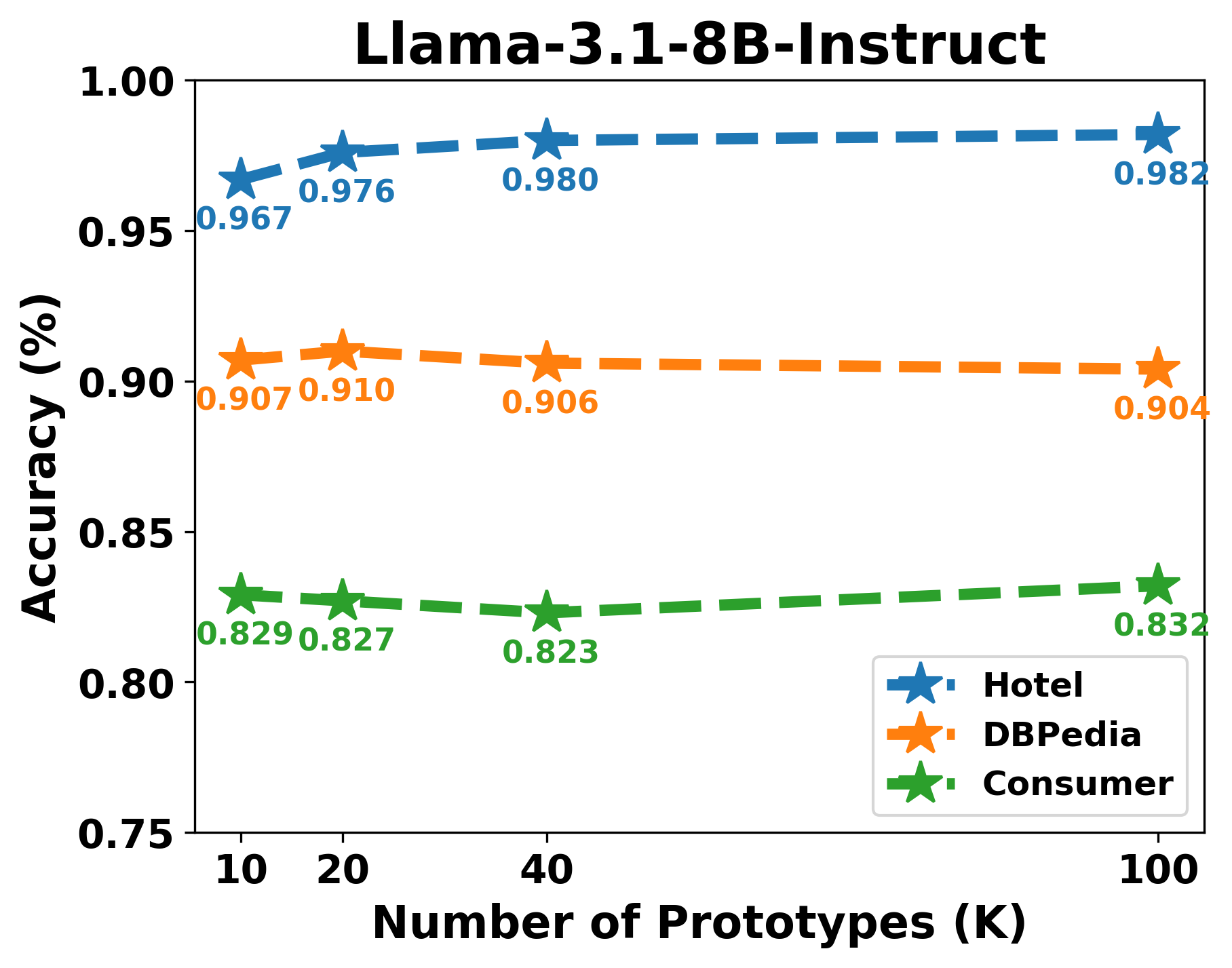}
    \caption{Llama-3.1-8B-Instruct}
  \end{subfigure}\hfill
  \begin{subfigure}[b]{0.5\textwidth}
    \centering
    \includegraphics[width=\textwidth]{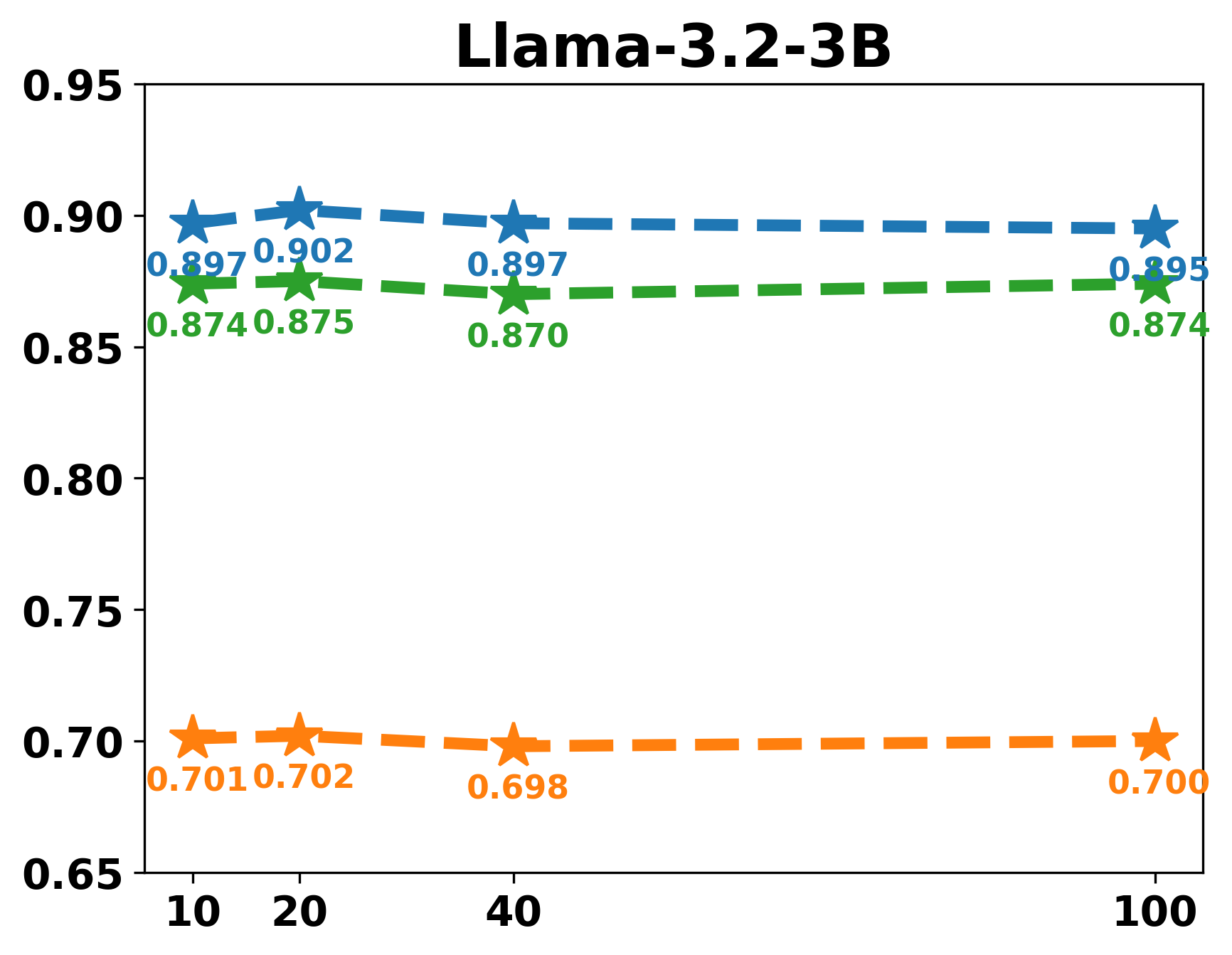}
    \caption{Llama-3.2-3B}
  \end{subfigure}

  \begin{subfigure}[b]{0.5\textwidth}
    \centering
    \includegraphics[width=\textwidth]{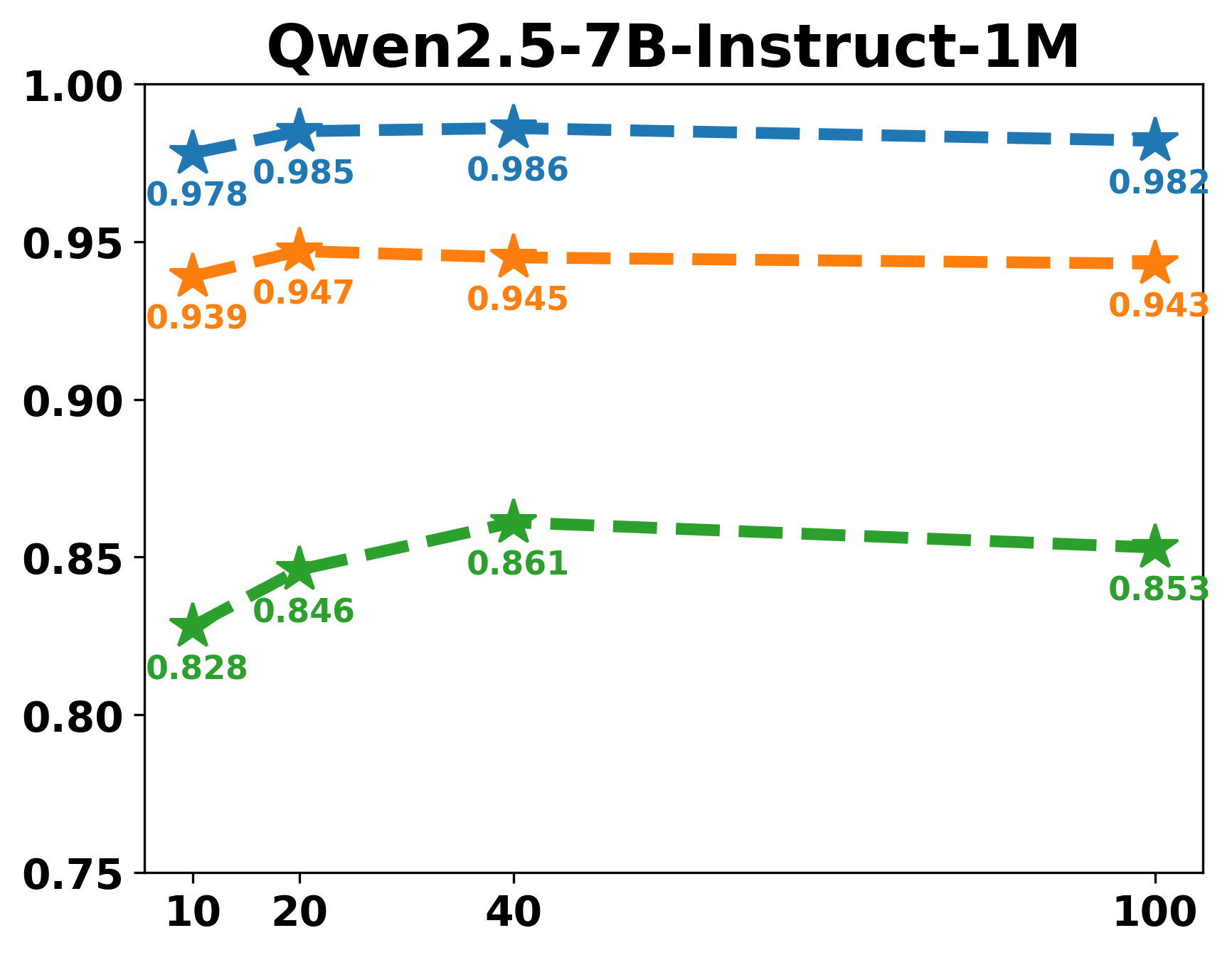}
    \caption{Qwen2.5‑7B‑Instruct‑1M}
  \end{subfigure}\hfill
  \begin{subfigure}[b]{0.5\textwidth}
    \centering
    \includegraphics[width=\textwidth]{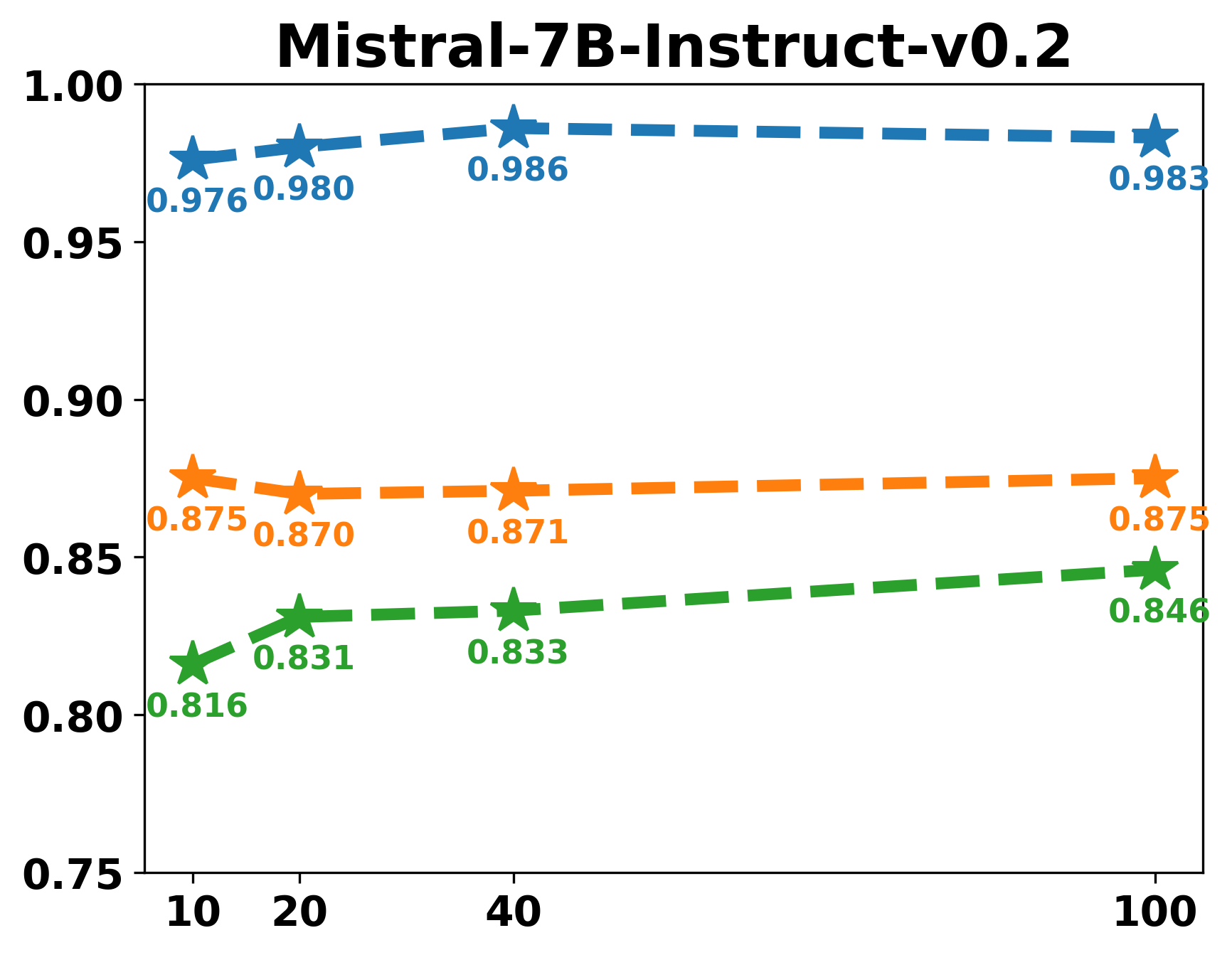}
    \caption{Mistral‑7B‑Instruct‑v0.2}
  \end{subfigure}

  \caption{Impact of the number of prototypes (\(K\)) on accuracy across different datasets and LLMs.
    Performance generally improves as \(K\) increases until reaching a plateau; the optimal \(K\) varies by dataset and model.}
    \label{fig:impact_prototypes}
\end{figure}

We investigate the effect of prototype count $K$ on ProtoSurE's classification performance. Figure~\ref{fig:impact_prototypes} present results for $K \in \{10,20,40,100\}$ across three datasets and four LLMs. Our findings demonstrate that accuracy consistently improves as $K$ increases from 10 to 40 before generally plateauing or showing diminishing returns. For Llama-3.1-8B, Hotel accuracy increases from 96.7\% at $K=10$ to 98.0\% at $K=40$, with only marginal improvement to 98.2\% at $K=100$. DBPedia performance with this model peaks earlier at $K=20$ (91.0\%), while Consumer accuracy continues improving to $K=100$ (83.2\%). Other LLMs exhibit similar patterns with optimal performance typically reached between $K=20$–40; notably, Qwen2.5-7B achieves its best results at $K=40$ across datasets. These findings suggest that $K=20$–40 offers an effective balance between model interpretability and representational capacity without excessive computational overhead.

\subsection{Case Study (RQ5)}
\begin{figure}[t]
\centering
\includegraphics[width=0.48\textwidth]{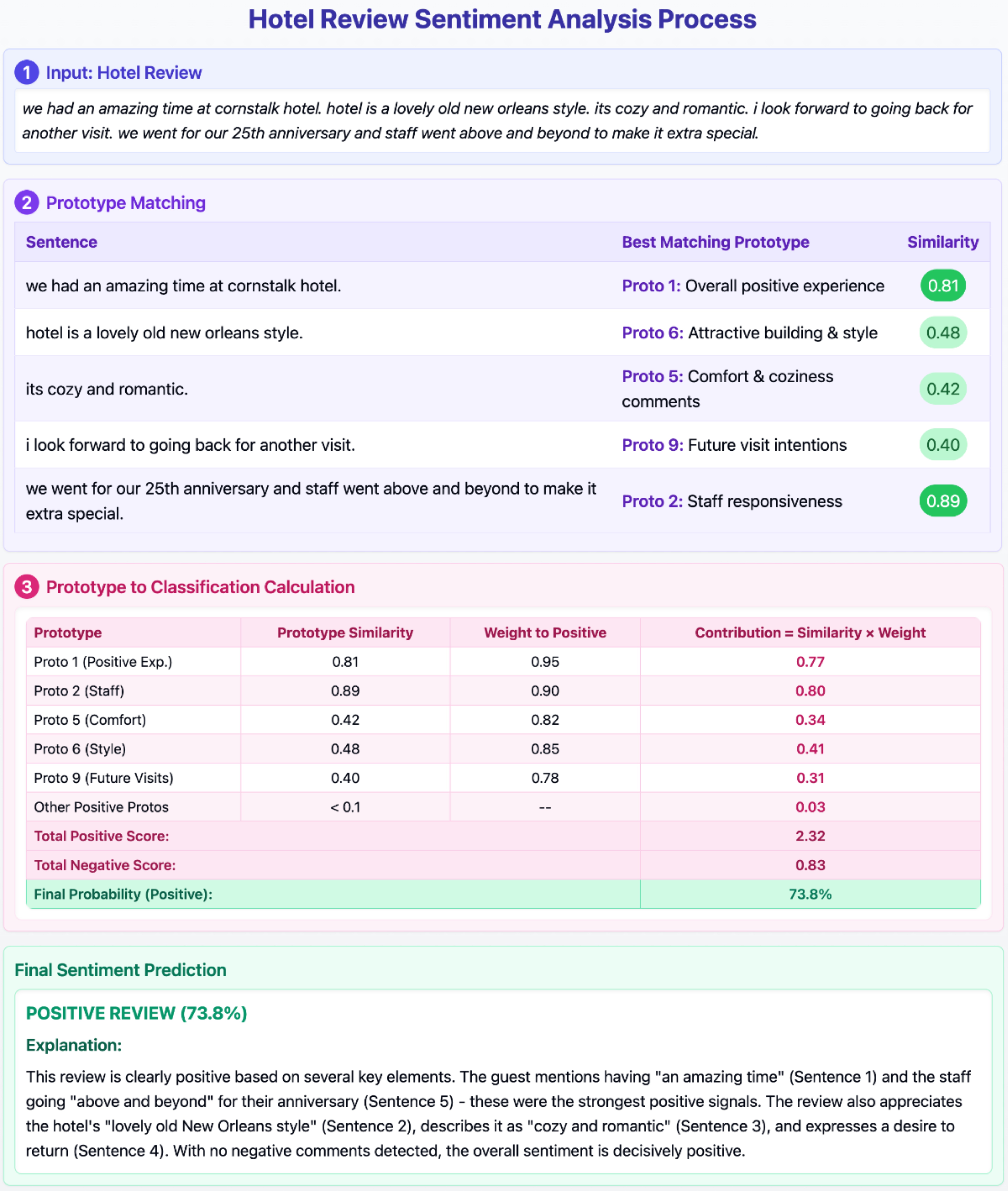}
\caption{Visualization of the sentiment analysis process for a hotel review using ProtoSurE. The visualization shows: (1) the original review text, (2) prototype matching between sentences and sentiment patterns, (3) aggregated similarity scores across prototype categories, and (4) final sentiment prediction with explanation.}
\label{fig:case_study}
\end{figure}

In this section, we present example visualizations of ProtoSurE . Figure~\ref{fig:case_study} demonstrates ProtoSurE's explainability through a hotel review example. The review contains five sentences that our system analyzes against ten prototype categories. Two sentences show strong matches: the mention of "an amazing time" (similarity score: 0.81) with Prototype 1 (Overall positive experience) and staff going "above and beyond" (similarity score: 0.89) with Prototype 2 (Staff responsiveness). Additionally, we observe substantial similarity between the other sentences and Prototypes 6 (Building \& style, 0.48), 5 (Comfort \& coziness, 0.42), and 9 (Future visit intentions, 0.40). When these sentence-level similarities are aggregated, they yield a decisive positive sentiment prediction (73.8\%), with negative prototype activations being negligible. This visualization demonstrates how ProtoSurE provides intuitive, interpretable reasoning for sentiment predictions by showing which specific textual elements activate meaningful prototype patterns. Additional examples are shown in Appendix~\ref{appendix:case}, Figure~\ref{fig:case_positive}, and Figure~\ref{fig:case_negative}.

\section{Conclusion}
In this work, we introduced \textbf{ProtoSurE}, a prototype‐based surrogate framework that delivers faithful and human‐understandable explanations for black‐box LLMs. By distilling LLM behavior into an interpretable model that matches sentences to semantically meaningful prototypes, ProtoSurE overcomes the limitations of existing post-hoc explanation approaches. Extensive experiments on four state‐of‐the‐art LLMs and four diverse datasets demonstrate that ProtoSurE faithfully reproduces LLM predictions (with an average fidelity over 88\%) while providing intuitive explanations. We also show that ProtoSurE is data‐efficient, requiring as few as 128 examples to approach full‐data performance. In future work, we aim to extend ProtoSurE to span‐level prototypes and explore its use in other modalities.

\section{Limitations}

\paragraph{Assumption of Sentence Coherence.}  
ProtoSurE assumes that each sentence in the input text conveys a coherent and self-contained semantic unit. This assumption may not hold in settings with irregular sentence boundaries, fragmented user inputs, or highly technical documents, where meaningful phrases span multiple sentences or clauses. As a result, some explanations may lose granularity or coherence in such contexts.

\paragraph{Simple Attribution Aggregation.}  
Our method aggregates sentence-level prototype contributions using simple averaging and scaling. Although this strategy is transparent and effective, it may fail to capture interactions between sentences or emphasize subtle dependencies. More advanced mechanisms (e.g., attention-based or learnable aggregation) could improve alignment fidelity but would introduce additional complexity.

\paragraph{Lack of Multilingual Support.}  
ProtoSurE is currently evaluated only on English-language data. While the framework itself is language-agnostic, sentence embeddings and attribution behavior can vary significantly across languages. Extending to multilingual or cross-lingual settings would require careful evaluation to ensure interpretability and alignment hold consistently.

\paragraph{No Modeling of Sentence Order.}  
Our approach treats sentences as an unordered set when aggregating prototype influences. For tasks that require reasoning over narrative structure or discourse flow, such as dialogue analysis or story understanding, this may limit the quality of the generated explanations. Incorporating temporal or structural modeling could enhance performance in such scenarios.

\section{Ethics}

This work focuses on improving transparency and trustworthiness of large language models through post-hoc explanation techniques. ProtoSurE is designed as an interpretability tool and does not modify the underlying LLM, thereby avoiding direct manipulation of model behavior. However, explanations produced by ProtoSurE are only as faithful as the surrogate model's approximation of the LLM, and misleading interpretations may arise if the surrogate fails to fully align. Care must be taken when deploying this method in high-stakes applications such as healthcare or legal decision-making. Additionally, while the datasets used in our experiments are publicly available and widely used for academic research, some may contain biased or sensitive content. We encourage practitioners to audit explanations for fairness and refrain from using ProtoSurE to justify harmful or discriminatory decisions.

\bibliographystyle{unsrtnat}
\bibliography{main}  

\clearpage
\appendix

\section{Datasets}
\label{appendix:datasets}
The IMDB dataset contains 25,000 balanced training and test samples and follows a binary sentiment classification format. The Hotel dataset includes 20,000 reviews evaluating 1,000 hotels.  Reviews with fewer than 10 characters or containing less than two sentences were excluded. 

The DBPedia dataset is a multiclass dataset extracted from Wikipedia. For the experiments in this paper, we use only 4 labels: ``Person,'' ``Animal,'' ``Building,'' and ``Natural Place.'' Similarly, the Consumer Complaints dataset is a multiclass dataset. For the experiments, we use only 4 classes: ``Checking or Savings Account,'' ``Credit Card or Prepaid Card,'' ``Debt Collection,'' and ``Mortgage.''

\section{Experimental Details}

\subsection{Baseline Methods}
\label{app:baselines}

We compare ProtoSurE against four widely-used post-hoc explanation methods:

\begin{itemize}
  \setlength\itemsep{0.5em}
  \item \textbf{SHAP} \cite{lundberg2017unified}: A unified approach for interpreting model predictions based on Shapley values, which attributes importance to individual tokens. SHAP assigns attribution scores by calculating the average contribution of each token to the prediction across all possible subsets of tokens. The key insight of SHAP is its ability to satisfy desirable properties like local accuracy, missingness, and consistency. For our implementation, we use KernelSHAP with 1000 samples per instance to approximate the Shapley values.
  
  \item \textbf{Integrated Gradients (IG)} \cite{sundararajan2017axiomatic}: A gradient-based feature attribution method that computes the integral of gradients with respect to inputs along a straight path from a baseline to the input. IG addresses the gradient saturation problem by integrating the gradients along this path. We use a zero embedding as the baseline and approximate the integral using 50 steps of the Riemann sum.
  
  \item \textbf{Occlusion (Occl)} \cite{zeiler2014visualizing}: A perturbation-based approach that measures feature importance by systematically masking individual tokens and observing the impact on the model's output. For each token, we replace it with a padding token and record the change in the output probability. The attribution score is proportional to the magnitude of this change. We experimented with different perturbation strategies (zero embedding, [MASK] token, random token) and found padding tokens to yield the most stable results.
  
  \item \textbf{DeepLIFT} \cite{shrikumar2017learning}: A backpropagation-based attribution technique that compares activation of each neuron to its reference activation and assigns contribution scores. DeepLIFT addresses the gradient saturation problem by considering the difference in activation from a reference state. We use a zero embedding as the reference input, following common practice in the literature.
\end{itemize}

Each explanation method is applied to explain predictions from four state-of-the-art LLMs:

\begin{itemize}
  \setlength\itemsep{0.5em}
  \item \textbf{Llama-3.1-8B-Instruct}: A larger instruction-tuned model from the Llama family with 8 billion parameters, designed to follow complex instructions.
  
  \item \textbf{Llama-3.2-3B}: A more compact model from the Llama family with 3 billion parameters, offering a balance between performance and computational efficiency.
  
  \item \textbf{Qwen2.5-7B-Instruct-1M}: An instruction-tuned multilingual model with 7 billion parameters, trained on diverse multilingual data.
  
  \item \textbf{Mistral-7B-Instruct-v0.2}: An instruction-tuned model from Mistral AI with 7 billion parameters, known for its strong performance on various NLP tasks.
\end{itemize}

For fair comparison, we adapt all baseline methods to provide sentence-level attributions by aggregating token-level scores within each sentence. Specifically, for each sentence, we compute the mean attribution score of all tokens in that sentence. We also experimented with alternative aggregation strategies (max, sum) but found mean aggregation to yield the most reasonable results.

\subsection{Evaluation Metrics}
\label{app:metrics}

We employ seven complementary metrics to assess the faithfulness of explanations:

\begin{itemize}
  \setlength\itemsep{0.5em}
  \item \textbf{Accuracy (Acc)}: Measures the percentage agreement between the surrogate model's predictions and the target LLM's predictions. Higher values indicate better fidelity of the explainer to the target model. Formally:
\begin{equation}
    \text{Acc} = \frac{1}{N}\sum_{i=1}^{N}\mathbb{1}[y_{\text{surr}}(x_i) = y_{\text{tgt}}(x_i)]
\end{equation}

where $y_{\text{surr}}(x_i) = \text{argmax}(f_{\text{surrogate}}(x_i))$ and $y_{\text{tgt}}(x_i) = \text{argmax}(f_{\text{target}}(x_i))$ are the predicted labels from the surrogate and target models, respectively.

  \item \textbf{Comprehensiveness (Comp)} \cite{DeYoung2020}: Quantifies how much the model's confidence drops when important sentences identified by the explanation method are removed. Calculated as:
  \begin{equation}
    \text{Comp}(x, e) = 1 - \frac{1}{L+1} \sum_{l=0}^{L} [f(x) - f(\tilde{x}^{(l)}_e)]
  \end{equation}
  where $f(x)$ is the model's confidence in the predicted class, and $\tilde{x}^{(l)}_e$ is the input with $l$ most important features (sentences in our case) removed. Higher values indicate that the removed content was truly important to the model's decision.
  
  \item \textbf{Sufficiency (Suff)} \cite{DeYoung2020}: Assesses whether the identified important sentences alone are sufficient to maintain the model's prediction. Calculated as:
  \begin{equation}
    \text{Suff}(x, e) = 1 - \frac{1}{L+1} \sum_{l=0}^{L} [f(x) - f(\hat{x}^{(l)}_e)]
  \end{equation}
  where $\hat{x}^{(l)}_e$ is the input with only the $l$ most important features present. Lower values indicate that the identified important sentences capture the essential information needed for the model's decision.
  
  \item \textbf{Decision Flip Fraction (DFF)} \cite{serrano2019attention}: Measures the fraction of feature removals needed to flip the decision:
  \begin{equation}
    \text{DFF}(x, e) = \frac{\arg\min_l \, g(\tilde{x}^{(l)}_e) \neq g(x)}{L}
  \end{equation}
  where $g(x)$ is the function that outputs the most likely class. We define DFF = 1 if no number of removals leads to a decision flip. Lower values are desirable, indicating that fewer important features need to be removed to change the model's decision.
  
  \item \textbf{Decision Flip with Most Important Sentence (DFS)} \cite{chrysostomou2021improving}: Measures whether removing just the single most important sentence changes the model's decision:
  \begin{equation}
    \text{DFS}(x, e) = \mathbb{1}_{g(\tilde{x}^{(1)}_e) \neq g(x)}
  \end{equation}
  where $\tilde{x}^{(1)}_e$ represents the input with the most important sentence removed, and $\mathbb{1}$ is the indicator function that equals 1 when the condition is true and 0 otherwise. Across a dataset, its average value gives the overall decision flip rate, and a higher value is desirable, indicating that the explanation correctly identifies sentences critical to the model's decision.
  
  \item \textbf{Deletion Rank Correlation (Del)} \cite{alvarez2018towards}: Evaluates the correlation between feature importance rankings and the impact of removing individual features:
  \begin{equation}
    \text{Del}(x, e) = \rho(\delta_f, e)
  \end{equation}
  where $\delta_f = [f(x) - f(x^{(1)}_{-,e}), \ldots, f(x) - f(x^{(L)}_{-,e})]$, with $x^{(l)}_{-,e}$ being the original input with only the $l$-th important feature removed, and $\rho$ is the Spearman rank correlation. Higher correlation suggests that suppressing more important features has a larger impact on the model prediction.
  
  \item \textbf{Insertion Rank Correlation (Ins)} \cite{luss2021leveraging}: Measures the correlation between feature importance rankings and the impact of sequentially adding features:
  \begin{equation}
    \text{Ins}(x, e) = \rho(v, [0, \ldots, L])
  \end{equation}
  where $v = [f(\tilde{x}^{(L)}_e), \ldots, f(\tilde{x}^{(0)}_e)]$, with $\tilde{x}^{(l)}_e$ being the sequence of inputs with increasingly more important features inserted. Higher correlation indicates better alignment between the explanation's importance rankings and the model's behavior when features are incrementally added.
\end{itemize}

\section{Target LLM Prompting}
\label{appendix:prompting}

To generate classifier predictions from the target LLMs, we use task-specific prompts designed to elicit consistent outputs. For binary classification tasks (IMDB and Hotel), we instruct the model to output either "A" (positive) or "B" (negative). For multi-class tasks (DBPedia and Consumer), we ask for an integer from 1 to 4. Table \ref{tab:prompts} shows the prompts used for each dataset and target LLM.

\begin{table}[t]
  \centering
  \caption{Prompts used for each dataset and target LLM. The placeholder \{review\} is replaced with the actual text to classify.}
  \label{tab:prompts}
  \scriptsize
  \begin{tabular}{p{2.2cm}|p{5.3cm}}
    \toprule
    \textbf{Dataset} & \textbf{Prompt Template} \\
    \midrule
    IMDB / Hotel & Classify the sentiment of the following review as either A (positive) or B (negative). Provide only the letter (A or B) as your response, with no additional explanation. Review: \{review\} Output: \\
    \midrule
    DBPedia & Classify the following Review into one of the categories: 1 (Person), 2 (Animal), 3 (Building), or 4 (Natural Place). Respond with only the corresponding integer (1, 2, 3, or 4) and no explanation. Your answer must be exactly one of: 1, 2, 3, or 4. Review: \{review\} Output: \\
    \midrule
    Consumer & Classify the following Review into one of the categories: 1 (Checking or Savings Account), 2 (Credit Card or Prepaid Card), 3 (Debt Collection), or 4 (Mortgage). Respond with only the corresponding integer (1, 2, 3, or 4) and no explanation. Your answer must be exactly one of: 1, 2, 3, or 4. Review: \{review\} Output: \\
    \bottomrule
  \end{tabular}
\end{table}

These prompt templates were consistently applied across all target LLMs (Llama-3.1-8B-Instruct, Llama-3.2-3B, Qwen2.5-7B-Instruct-1M, and Mistral-7B-Instruct-v0.2). The instructions emphasize producing only the classification label without additional explanation, which ensures consistent outputs for our evaluation. For our surrogate model training, we collected 2,000 labeled examples for each dataset-model combination using these prompts.

\section{Faithfulness Evaluation}
\label{appendix:faith_eval}
\begin{table*}[t]
 \centering
 \caption{Faithfulness evaluation : Faithfulness metrics across four target LLMs and four datasets. We report  Accuracy (Acc, \% $\uparrow$), Comprehensiveness (Comp $\uparrow$), Sufficiency (Suff $\downarrow$), Decision Flip Fraction (DFF, \% $\downarrow$), Decision Flip with Most Important Sentence (DFS, \% $\uparrow$), Deletion Rank Correlation (Del $\uparrow$), and Insertion Rank Correlation (Ins $\uparrow$). Arrows indicate whether higher ($\uparrow$) or lower ($\downarrow$) values are better. Avg represents the average value across all datasets and LLMs, while Avg Rank shows the average ranking among all methods (lower is better). Best results for each metric are in \textbf{bold}.}
 \label{tab:faithfulness_part2}
 \tiny
 \setlength{\tabcolsep}{1.8pt}
 \renewcommand{\arraystretch}{1.05}
 \begin{tabular}{ll|cccc|cccc|cccc|cccc|c|c}
   \toprule
   & & \multicolumn{4}{c|}{\textbf{Llama-3.1-8B}} & \multicolumn{4}{c|}{\textbf{Llama-3.2-3B}} & \multicolumn{4}{c|}{\textbf{Qwen2.5-7B}} & \multicolumn{4}{c|}{\textbf{Mistral-7B}} & \multirow{2}{*}{\textbf{Avg}} & \multirow{2}{*}{\textbf{Avg Rank}} \\
   \cmidrule{3-18}
   \textbf{Metric} & \textbf{Method} & IMDB & Hotel & DBPedia & Consumer & IMDB & Hotel & DBPedia & Consumer & IMDB & Hotel & DBPedia & Consumer & IMDB & Hotel & DBPedia & Consumer & & \\
      \midrule
   \multirow{1}{*}{Acc (\%) $\uparrow$} 
   & \textbf{ProtoSurE} & \textbf{93.5} & \textbf{98.4} & \textbf{91.0} & \textbf{83.2} & \textbf{85.5} & \textbf{90.7} & \textbf{70.2} & \textbf{80.1} & \textbf{95.6} & \textbf{98.6} & \textbf{94.7} & \textbf{87.0} & \textbf{94.7} & \textbf{98.7} & \textbf{87.0} & \textbf{84.4} & \textbf{89.58} & \textbf{--} \\
   \midrule
   \multirow{5}{*}{Comp $\uparrow$} 
   & SHAP      & 0.105 & 0.088 & 0.176 & 0.231 & -0.006 & 0.038 & 0.152 & 0.042 & 0.112 & 0.095 & 0.231 & 0.268 & 0.159 & 0.148 & 0.234 & 0.291 & 0.153 & 3.50 \\
   & IG        & 0.118 & 0.086 & 0.188 & 0.241 & -0.006 & 0.083 & 0.161 & 0.048 & 0.121 & 0.108 & 0.237 & 0.273 & 0.162 & 0.146 & 0.250 & 0.308 & 0.164 & 2.19 \\
   & Occl      & 0.106 & 0.086 & 0.179 & 0.231 & -0.005 & 0.064 & 0.159 & 0.041 & 0.108 & 0.090 & 0.231 & 0.271 & 0.154 & 0.146 & 0.243 & 0.302 & 0.155 & 3.44 \\
   & DeepLIFT  & 0.092 & 0.067 & 0.167 & 0.219 & -0.008 & 0.070 & 0.144 & 0.039 & 0.094 & 0.082 & 0.198 & 0.208 & 0.103 & 0.098 & 0.204 & 0.216 & 0.129 & 4.88 \\
   & \textbf{ProtoSurE} & \textbf{0.156} & \textbf{0.141} & \textbf{0.271} & \textbf{0.268} & \textbf{0.002} & \textbf{0.109} & \textbf{0.212} & \textbf{0.104} & \textbf{0.172} & \textbf{0.146} & \textbf{0.283} & \textbf{0.317} & \textbf{0.365} & \textbf{0.343} & \textbf{0.389} & \textbf{0.351} & \textbf{0.235} & \textbf{1.00} \\

   \midrule
   \multirow{5}{*}{Suff $\downarrow$} 
   & SHAP      & \textbf{0.205} & 0.192 & 0.234 & 0.257 & 0.063 & 0.050 & 0.173 & 0.027 & 0.196 & 0.185 & 0.171 & 0.190 & 0.159 & 0.137 & 0.184 & 0.215 & 0.165 & 4.03 \\
   & IG        & 0.236 & 0.114 & 0.218 & 0.244 & 0.104 & 0.094 & 0.160 & \textbf{0.022} & 0.138 & 0.126 & \textbf{0.166} & 0.186 & 0.098 & 0.086 & \textbf{0.164} & \textbf{0.201} & 0.141 & 2.09 \\
   & Occl      & 0.208 & 0.119 & 0.229 & 0.253 & 0.097 & 0.088 & 0.164 & 0.026 & 0.145 & 0.132 & 0.171 & \textbf{0.184} & \textbf{0.095} & 0.086 & 0.174 & 0.207 & 0.144 & 2.78 \\
   & DeepLIFT  & 0.234 & 0.126 & 0.240 & 0.261 & 0.096 & 0.084 & 0.181 & 0.026 & 0.157 & 0.140 & 0.203 & 0.244 & 0.175 & 0.157 & 0.216 & 0.287 & 0.172 & 4.41 \\
   & \textbf{ProtoSurE} & 0.214 & \textbf{0.060} & \textbf{0.207} & \textbf{0.231} & \textbf{0.051} & \textbf{0.039} & \textbf{0.156} & 0.024 & \textbf{0.116} & \textbf{0.102} & 0.170 & 0.189 & 0.152 & \textbf{0.136} & 0.173 & 0.210 & \textbf{0.131} & \textbf{1.69} \\
   \midrule
   \multirow{5}{*}{DFF (\%) $\downarrow$} 
   & SHAP      & 0.710 & 0.694 & 0.859 & 0.699 & 0.695 & 0.681 & 0.723 & 0.759 & 0.712 & 0.703 & 0.860 & 0.799 & 0.534 & 0.523 & 0.849 & 0.752 & 0.722 & 3.62 \\
   & IG        & 0.684 & 0.694 & 0.841 & 0.678 & 0.662 & 0.653 & \textbf{0.698} & 0.730 & 0.695 & 0.689 & 0.854 & 0.791 & 0.541 & 0.530 & 0.819 & 0.730 & 0.706 & 2.28 \\
   & Occl      & 0.709 & 0.702 & 0.858 & 0.707 & 0.671 & 0.658 & 0.721 & 0.803 & 0.701 & 0.694 & 0.867 & 0.798 & 0.545 & 0.538 & 0.836 & 0.748 & 0.722 & 3.69 \\
   & DeepLIFT  & 0.716 & 0.704 & 0.877 & 0.720 & 0.667 & 0.652 & 0.723 & 0.800 & 0.723 & 0.710 & 0.918 & 0.881 & 0.538 & 0.526 & 0.900 & 0.865 & 0.745 & 4.34 \\
   & \textbf{ProtoSurE} & \textbf{0.645} & \textbf{0.634} & \textbf{0.837} & \textbf{0.633} & \textbf{0.641} & \textbf{0.637} & 0.719 & \textbf{0.721} & \textbf{0.662} & \textbf{0.651} & \textbf{0.850} & \textbf{0.771} & \textbf{0.512} & \textbf{0.504} & \textbf{0.815} & \textbf{0.728} & \textbf{0.685} & \textbf{1.06} \\
   \midrule
   \multirow{5}{*}{DFS (\%) $\uparrow$} 
   & SHAP      & 0.170 & 0.180 & \textbf{0.111} & 0.212 & 0.045 & 0.041 & 0.250 & 0.134 & 0.023 & 0.020 & 0.121 & 0.081 & 0.290 & 0.300 & 0.090 & 0.081 & 0.134 & 3.56 \\
   & IG        & 0.196 & 0.205 & 0.091 & 0.202 & 0.043 & 0.039 & 0.294 & 0.175 & 0.035 & 0.031 & 0.162 & 0.110 & 0.285 & 0.280 & \textbf{0.162} & \textbf{0.222} & 0.158 & 2.81 \\
   & Occl      & 0.197 & 0.205 & 0.091 & 0.202 & 0.037 & 0.031 & 0.266 & 0.130 & 0.037 & 0.033 & \textbf{0.162} & \textbf{0.111} & 0.287 & 0.281 & 0.159 & 0.222 & 0.153 & 2.88 \\
   & DeepLIFT  & 0.183 & 0.190 & 0.101 & 0.182 & 0.037 & 0.033 & 0.255 & 0.128 & 0.038 & 0.033 & 0.081 & 0.101 & 0.268 & 0.274 & 0.091 & 0.051 & 0.128 & 4.06 \\
   & \textbf{ProtoSurE} & \textbf{0.225} & \textbf{0.216} & 0.091 & \textbf{0.221} & \textbf{0.187} & \textbf{0.180} & \textbf{0.310} & \textbf{0.192} & \textbf{0.051} & \textbf{0.045} & 0.155 & 0.103 & \textbf{0.325} & \textbf{0.315} & 0.152 & 0.213 & \textbf{0.187} & \textbf{1.69} \\

   \midrule
   \multirow{5}{*}{Del $\uparrow$} 
   & SHAP      & \textbf{-0.005} & 0.003 & 0.137 & 0.106 & 0.015 & 0.030 & 0.145 & 0.027 & 0.012 & 0.015 & 0.163 & 0.053 & -0.021 & -0.017 & 0.138 & 0.039 & 0.053 & 4.00 \\
   & IG        & -0.028 & -0.020 & 0.192 & 0.166 & 0.003 & 0.039 & 0.173 & 0.080 & 0.021 & 0.018 & 0.194 & 0.076 & -0.017 & -0.011 & 0.168 & 0.112 & 0.075 & 3.09 \\
   & Occl      & -0.031 & -0.023 & 0.216 & 0.183 & \textbf{0.056} & 0.056 & 0.196 & 0.098 & \textbf{0.034} & \textbf{0.028} & \textbf{0.216} & \textbf{0.169} & -0.016 & -0.011 & \textbf{0.201} & \textbf{0.162} & 0.095 & 2.03 \\
   & DeepLIFT  & -0.019 & -0.011 & 0.165 & 0.091 & 0.003 & 0.012 & 0.135 & 0.062 & -0.011 & -0.008 & -0.051 & -0.068 & 0.009 & 0.013 & 0.146 & -0.099 & 0.023 & 4.12 \\
   & \textbf{ProtoSurE} & -0.035 & \textbf{0.084} & \textbf{0.237} & \textbf{0.187} & 0.020 & \textbf{0.073} & \textbf{0.204} & \textbf{0.112} & 0.031 & 0.027 & 0.210 & 0.153 & \textbf{0.022} & \textbf{0.017} & 0.192 & 0.143 & \textbf{0.105} & \textbf{1.75} \\
   \midrule
   \multirow{5}{*}{Ins $\uparrow$} 
   & SHAP      & 0.161 & 0.153 & 0.416 & 0.354 & -0.058 & 0.203 & 0.399 & 0.448 & 0.201 & 0.192 & \textbf{0.645} & 0.583 & 0.184 & 0.179 & 0.447 & 0.430 & 0.325 & 2.88 \\
   & IG        & 0.219 & \textbf{0.210} & 0.413 & 0.329 & -0.043 & 0.230 & 0.392 & 0.439 & \textbf{0.212} & \textbf{0.204} & 0.638 & \textbf{0.586} & 0.183 & 0.178 & 0.436 & \textbf{0.432} & 0.334 & 2.28 \\
   & Occl      & \textbf{0.220} & 0.210 & 0.389 & 0.316 & -0.043 & \textbf{0.233} & 0.385 & 0.405 & 0.207 & 0.198 & 0.636 & 0.567 & 0.182 & 0.176 & 0.408 & 0.400 & 0.324 & 3.22 \\
   & DeepLIFT  & 0.164 & 0.157 & 0.380 & 0.322 & -0.049 & 0.219 & 0.397 & 0.420 & 0.186 & 0.180 & 0.590 & 0.523 & 0.162 & 0.156 & 0.400 & 0.263 & 0.297 & 4.50 \\
   & ProtoSurE & 0.215 & 0.205 & \textbf{0.422} & \textbf{0.342} & \textbf{-0.035} & 0.223 & \textbf{0.405} & \textbf{0.450} & 0.202 & 0.196 & 0.640 & 0.580 & \textbf{0.202} & \textbf{0.195} & \textbf{0.450} & 0.425 & \textbf{0.336} & \textbf{2.12} \\
   \bottomrule
 \end{tabular}
\end{table*}

Table~\ref{tab:faithfulness_part2} presents a detailed comparison of ProtoSurE and four token-level attribution baselines—SHAP, Integrated Gradients (IG), Occlusion, and DeepLIFT—across four datasets and four target LLMs. We evaluate using multiple faithfulness metrics: Accuracy (Acc), Comprehensiveness (Comp), Sufficiency (Suff), Decision Flip Fraction (DFF), Decision Flip with Most Important Sentence (DFS), Deletion Rank Correlation (Del), and Insertion Rank Correlation (Ins).

ProtoSurE consistently outperforms all baselines across most metrics. It achieves the highest Comprehensiveness score (0.235 vs. IG's 0.164), indicating superior ability to identify truly influential input components. It also yields the lowest Sufficiency score (0.131 vs. IG's 0.141), demonstrating that the remaining tokens after removing important ones are less sufficient for prediction—suggesting better focus on salient content. ProtoSurE also shows lower DFF (0.685 vs. IG's 0.706), suggesting higher stability, and higher DFS (0.187 vs. IG's 0.158), reflecting better localization of critical evidence. Additionally, it leads in Del (0.105 vs. Occlusion's 0.095) and Ins (0.336 vs. IG's 0.334), reinforcing that the ranked importance aligns more faithfully with model behavior under perturbations.

Overall, ProtoSurE ranks first across nearly all metrics, validating its effectiveness as a faithful and interpretable explanation framework, particularly when explaining black-box LLM predictions using sentence-level semantic abstractions.

\section{Encoder Impact}
\label{appendix: encoder}
\begin{table}[t]
\centering
\caption{Impact of different encoders on ProtoSurE's Accuracy (\%) across all target LLMs and datasets. Best results are in \textbf{bold}.}
\label{tab:embedding_impact}
\small
\setlength{\tabcolsep}{2pt}
\renewcommand{\arraystretch}{1.1}
\begin{tabular}{lcccc}
\toprule
\textbf{Encoder} & \textbf{Hotel} & \textbf{DBPedia} & \textbf{Consumer} & \textbf{Avg Rank}  \\
\midrule
\multicolumn{5}{c}{\textit{Llama-3.1-8B-Instruct}} \\
\midrule
SBERT       & 0.970  & 0.907 & 0.859  & 3.33 \\
BGE       & \textbf{0.991}  & 0.906 & \textbf{0.863}  & 2.00 \\
GTE                & 0.984  & \textbf{0.910} & 0.862 & \textbf{1.67} \\
E5   & 0.989  & 0.908 & 0.860 & 2.67 \\
T5        & 0.978  & 0.891 & 0.844 & 5.00 \\
\midrule
\multicolumn{5}{c}{\textit{Llama-3.2-3B}} \\
\midrule
SBERT       & 0.867 & 0.712 & 0.859  & 4.33 \\
BGE       & 0.901 & 0.708 & 0.873  & 3.00 \\
GTE                & \textbf{0.907} & 0.702 & 0.874 & 2.33 \\
E5   & 0.903 & \textbf{0.717} & 0.876 & \textbf{2.00} \\
T5        & 0.892 & 0.702 & \textbf{0.881} & 3.33 \\
\midrule
\multicolumn{5}{c}{\textit{Qwen2.5-7B-Instruct-1M}} \\
\midrule
SBERT       & 0.976 & 0.946 & 0.864  & 3.33 \\
BGE       & \textbf{0.990} & 0.943 & 0.859  & 2.67 \\
GTE                & 0.986 & \textbf{0.947} & \textbf{0.870} & \textbf{1.33} \\
E5   & \textbf{0.990} & 0.945 & 0.846 & 2.67 \\
T5        & 0.982 & 0.936 & 0.861 & 4.00 \\
\midrule
\multicolumn{5}{c}{\textit{Mistral-7B-Instruct-v0.2}} \\
\midrule
SBERT       & 0.981 & \textbf{0.875} & 0.841  & 3.00 \\
BGE       & 0.986 & 0.873 & 0.834  & 3.33 \\
GTE                & 0.987 & 0.870 & 0.844  & 2.67 \\
E5   & \textbf{0.990} & 0.861 & 0.835 & \textbf{2.33} \\
T5        & 0.986 & 0.870 & \textbf{0.846} & 3.67 \\

\bottomrule
\end{tabular}
\end{table}
We investigate the effect of different sentence encoders on ProtoSurE's classification performance across three datasets (Hotel, DBPedia, Consumer) and four target LLMs. As shown in Table~\ref{tab:embedding_impact}, we compare five commonly used sentence encoders: SBERT, BGE, GTE, E5, and T5.

GTE consistently achieves strong performance, ranking the best on average across most settings. For instance, it achieves the highest overall accuracy on the DBPedia and Consumer datasets under Llama-3.1-8B and Qwen2.5-7B. BGE and E5 also perform competitively across all LLMs. T5, by contrast, generally lags behind in both accuracy and average rank. These results suggest that selecting a high-quality encoder, particularly one tuned for semantic similarity like GTE or BGE, can substantially improve ProtoSurE's alignment with the underlying LLM's decision boundary.

\section{Case Study}
\label{appendix:case}

\begin{figure}[t]

\centering
\includegraphics[width=0.48\textwidth]{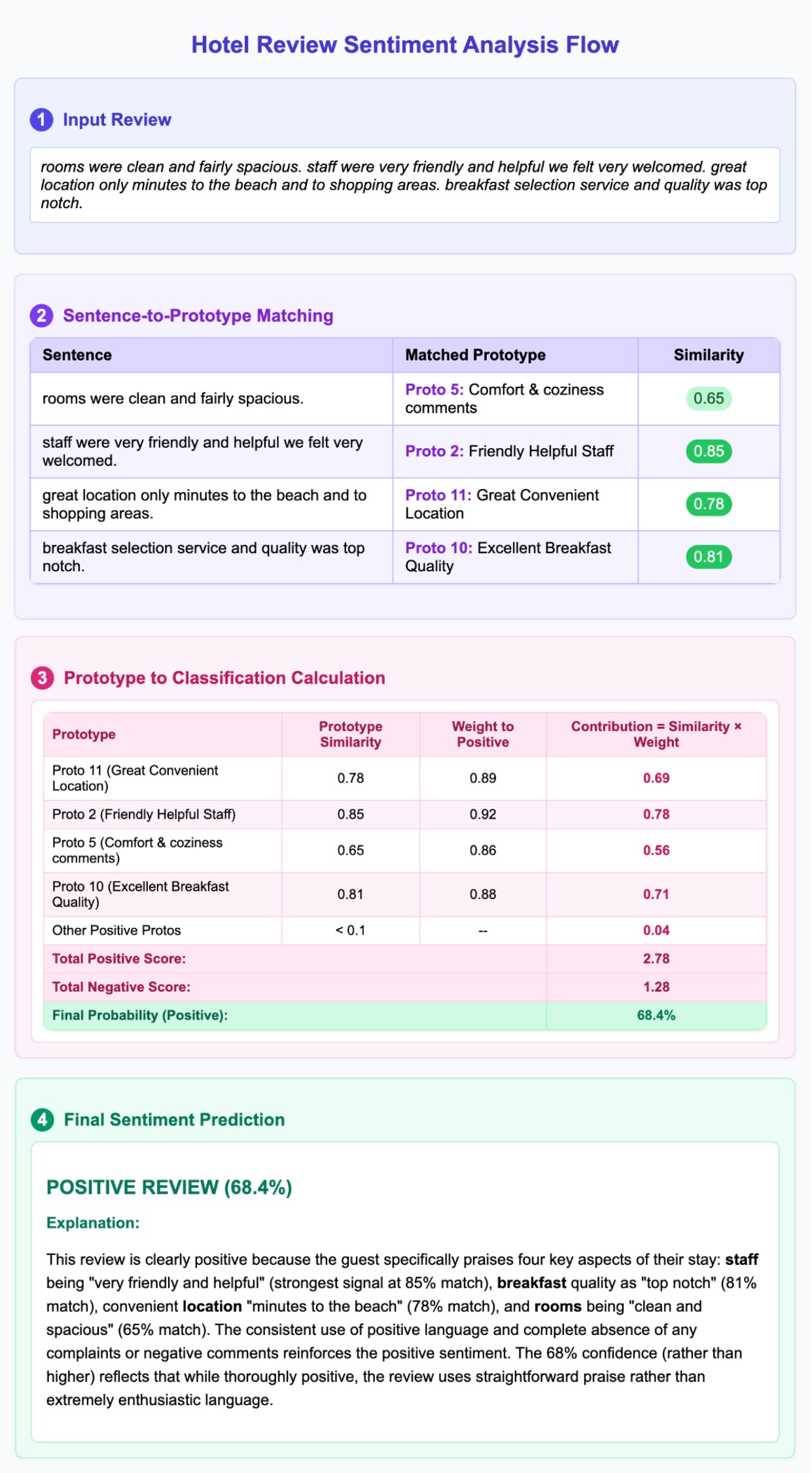}
\caption{A positive review correctly classified by \textsc{ProtoSurE}, with aligned prototypes highlighting praise for staff, room, location, and breakfast.}
\label{fig:case_positive}
\end{figure}

\begin{figure}[t]

\centering
\includegraphics[width=0.48\textwidth]{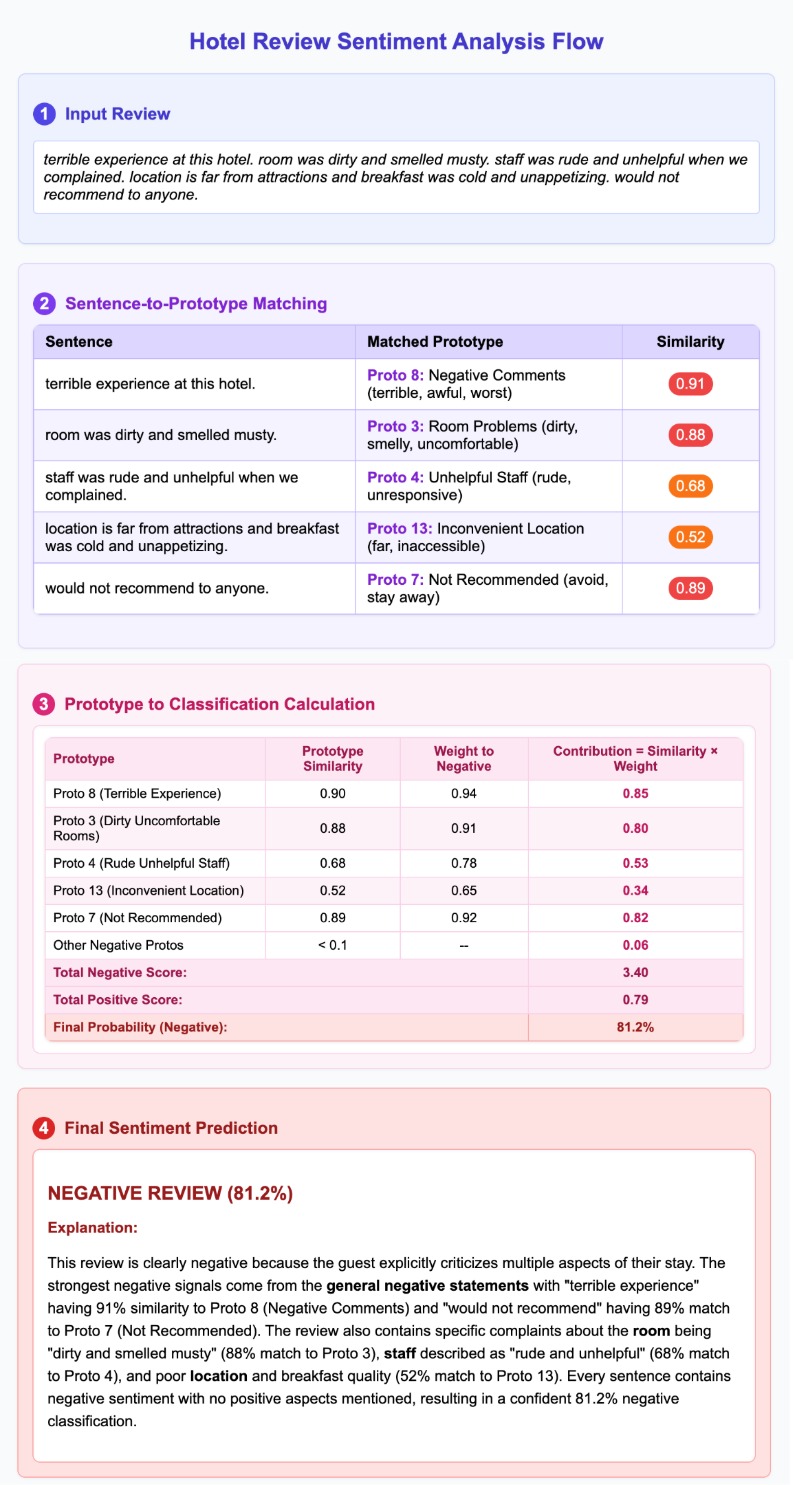}
\caption{A negative review where \textsc{ProtoSurE} identifies strong alignment with negative prototypes such as room complaints and general dissatisfaction.}
\label{fig:case_negative}
\end{figure}

We provide qualitative examples to illustrate how \textsc{ProtoSurE} decomposes predictions into interpretable prototype-level contributions. Figures~\ref{fig:case_positive} and~\ref{fig:case_negative} present two hotel reviews—one positive and one negative—and show how each sentence is matched to a semantically aligned prototype, contributing to the final classification.

In the positive case (Figure~\ref{fig:case_positive}), the model identifies four aspects with high similarity to positive prototypes: room quality (Proto 6), staff friendliness (Proto 4), location (Proto 0), and breakfast quality (Proto 9). All sentences match prototypes with high confidence scores (0.65–0.85), leading to a strong positive classification with 68.4\% probability.

Conversely, the negative review (Figure~\ref{fig:case_negative}) expresses dissatisfaction across multiple dimensions. Sentences match closely with prototypes representing general negative sentiment (Proto 8), room complaints (Proto 3), and staff issues (Proto 4). Each prototype receives a high similarity score (up to 0.91), with negative prototype activations outweighing the positive ones and yielding a 76.5\% confidence for the negative class.

These examples demonstrate \textsc{ProtoSurE}'s ability to provide faithful, aspect-aware explanations by aligning input sentences with relevant prototypes and transparently combining their contributions into the final decision.

\end{document}